\documentclass[letterpaper, 10 pt, conference]{ieeeconf}  % Comment this line out if you need a4paper

\IEEEoverridecommandlockouts                              % This command is only needed if 
\usepackage{amsmath} % assumes amsmath package installed
\usepackage{amssymb}  % assumes amsmath package installed
\usepackage{bm}
\usepackage{graphicx}
\usepackage{caption}
\usepackage[table]{xcolor}
\usepackage{subcaption}
\usepackage{xspace}
\usepackage{multirow}
\usepackage{multicol}
\let\labelindent\relax
\usepackage[inline,shortlabels]{enumitem}
\usepackage[bookmarks = true,
            colorlinks = true,
            linkcolor = black,
            urlcolor  = black,
            citecolor = black,
            anchorcolor = black]{hyperref}
\usepackage[ruled,vlined,linesnumbered]{algorithm2e}

\newcommand{\sref}[1]{Section~\ref{#1}}
\newcommand{\eref}[1]{Equation~\ref{#1}}
\newcommand{\fref}[1]{Figure~\ref{#1}}
\newcommand{\tref}[1]{Table~\ref{#1}}

\newcommand{\true}{\textrm{T}}
\newcommand{\false}{\textrm{F}}
\newcommand{\bools}{\{\true,\false\}}

\newcommand{\sar}{\textit{safety-aware reasoning}\xspace}
\newcommand{\Sar}{\textit{Safety-aware reasoning}\xspace}

\newcommand{\sa}{\textit{safety-aware}\xspace}
\newcommand{\hrrt}{\textit{human-robot red team}\xspace}
\newcommand{\HRRT}{\textit{HRRT}\xspace}
\newcommand{\HRRTing}{\textit{human-robot red teaming}\xspace}
\newcommand{\HRRTed}{\textit{human-robot red teamed}\xspace}
\newcommand{\hrrts}{\textit{human-robot red teams}\xspace}
\newcommand{\HRRTs}{\textit{HRRTs}\xspace}
\newcommand{\rmp}{\textit{risk mitigating policy}\xspace}

\newcommand{\rmau}{\textit{risk mitigating action-utility}\xspace}

\newcommand{\rmaum}{\textit{risk mitigating action-utility model}\xspace}
\newcommand{\rmaums}{\textit{risk mitigating action-utility models}\xspace}

\title{\LARGE \bf
Human-Robot Red Teaming for Safety-Aware Reasoning %\\
}

\author{Emily Sheetz\textsuperscript{1,2}, Emma Zemler\textsuperscript{2}, Misha Savchenko\textsuperscript{2}, Connor Rainen\textsuperscript{2}, Erik Holum\textsuperscript{2}, \\
Jodi Graf\textsuperscript{2}, Andrew Albright\textsuperscript{2},
Shaun Azimi\textsuperscript{2}, and Benjamin Kuipers\textsuperscript{1}%
\thanks{Authors with \textsuperscript{1}University of Michigan and \textsuperscript{2}NASA Johnson Space Center.
}%
\thanks{Disclaimer: Trade names and trademarks are used in this report for identification only.  Their usage does not constitute an official endorsement, either expressed or implied, by the National Aeronautics and Space Administration (NASA).}%
}

\begin{document}

\maketitle
\thispagestyle{empty}
\pagestyle{empty}

%%%%%%%%%%%%%%%%%%%%%%%%%%%%%%%%%%%%%%%%%%%%%%%%%%%%%%%%%%%%%%%%%%%%%%%%%%%%%%%%
\begin{abstract}

While much research explores improving robot capabilities, there is a deficit in researching how robots are expected to perform tasks safely, especially in high-risk problem domains.
Robots must earn the trust of human operators in order to be effective collaborators in safety-critical tasks, specifically those where robots operate in human environments. % alongside humans.
We propose the \HRRTing paradigm for \sar.
We expect humans and robots to work together to challenge assumptions about an environment and explore the space of hazards that may arise.
This exploration will enable robots to perform \sar, specifically hazard identification, risk assessment, risk mitigation, and safety reporting.
% In this paper, we explore how the \hrrt can teach robots to appropriately assess risks.
We demonstrate that:
\begin{enumerate*}[label=(\alph*)]
    \item \HRRTing allows human-robot teams to plan to perform tasks safely in a variety of domains, and
    \item robots with different embodiments can learn to operate safely in two different environments---a lunar habitat and a household---with varying definitions of safety.
\end{enumerate*}
Taken together, our work on \HRRTing for \sar demonstrates the feasibility of this approach for safely operating and promoting trust on human-robot teams in safety-critical problem domains.

\end{abstract}

%%%%%%%%%%%%%%%%%%%%%%%%%%%%%%%%%%%%%%%%%%%%%%%%%%%%%%%%%%%%%%%%%%%%%%%%%%%%%%%%

\section{INTRODUCTION}
\label{sec:intro}

% \todo{remove NL/dialogue entirely to clarify the point?}

\begin{figure}[t!]
    \centering
    \includegraphics[width=0.48\textwidth,keepaspectratio]{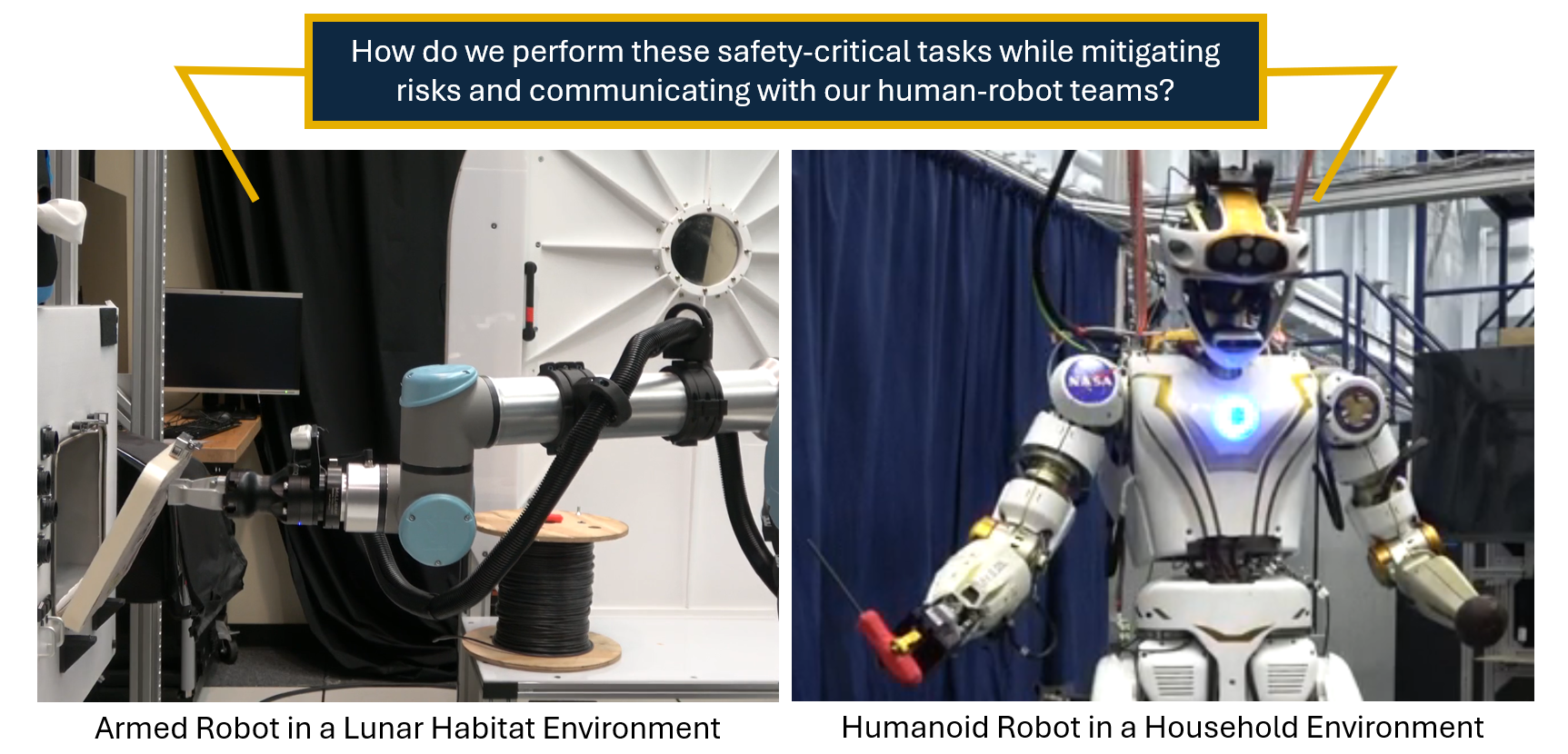}
    \caption{Robots with different embodiments acting in different environments must be able to reason over the safety of a task, mitigate risks, and report their assessments to other agents on human-robot teams.
    }
    \label{fig:teaser}
\end{figure}

% value proposition, problem, motivation (if we solve x, we get y)
Enabling robots to reason over risks is a crucial capability of performing collaborative assistive tasks in safety-critical domains.
A key aspect of safety is appropriate trust between robot and human operator, which can be earned through clear communication and explainable robot behavior.
In particular, we want
% to allow users to use intuitive modes to communicate task goals and 
to ensure robots can assess risks and communicate safety issues to human operators, as depicted in \fref{fig:teaser}.
% As robots become increasingly capable of performing tasks in different domains,
% Natural language has gained much interest as an intuitive interfaces for users to interact with robots, especially for non-expert users without programming experience.
% Due to the prevalence of language for interpersonal communication,
% language-based interactions also allow more non-expert users to intuitively interact with robots in many diverse domains.
It is imperative that robots reason over task
safety %natural language commands within safety-critical domains
and report their risk assessments
% provide clear explanations for their behavior 
in order to earn operator trust.

% challenge and why problem has not been solved yet (x is hard, why is it hard?)
% Reasoning over natural language commands in a safety-aware manner is particularly challenging due to the abundant sensory signals language supports.
There is a consensus that robot safety is important, especially in domains where humans and robots operate in the same environment.
Despite broad agreement on the importance of safety,
% natural language 
existing systems often fail to consider
% research often does not consider 
how robots should execute commands safely \cite{Zhang2022}, instead overtrusting human operators to evaluate safety \cite{Lee2004, Robinette2016}.
Furthermore, safety issues are likely to arise when an agent's simplifying model of the unboundedly complex world does not include details that prove to be critical. Poorly constructed simplifying models can result in disastrous consequences.
For safety-critical domains, it is important to have an adequately complex model of the world, identify what is left out of the current model, and account for unmodeled events.
% for safety, the question may not be "what do we model?" but more importantly "what has been left out of our current model, so we can account for these things?"

% insight (if we think about x in this way, we get this new insight to approach the problem in this way and offer a solution)
To address the challenges of \sar, % over natural language commands,
we take inspiration from literature on trust and on red teaming.
In cooperative tasks, trust allows agents to make simplifying assumptions about other agents' behaviors \cite{Kuipers2022}.
But poorly calibrated trust \cite{Lee2004, Robinette2016} can be dangerous and make robot operations unsafe.
In order for robots to earn the trust of their human operators, we expect robots to provide clear explanations about their risk assessments and behaviors in safety-critical tasks.
Red teaming strategies help identify vulnerabilities and strengthen weaknesses in models.
Through red teaming, the robot can ensure its model sufficiently captures the risks that may arise in a safety-critical task.
% safety looks different in different tasks, so formulation must be generic enough to support different definitions of safety; checks and balances between robot and operator, neither should be entirely out of the loop, agreed upon definition of safety

% contributions (in this paper, we take this approach to solving x, evaluating our approach, and explaining how much we achieve benefit y)
We propose a \textbf{\HRRTing} paradigm to allow robots to perform \textbf{\sar}.
We expect robots operating on human-robot teams to understand the complexity of acting safely in a problem domain, identify hazards, assess risks, mitigate risks, and report on safety.
In this paper, we explore how the \hrrt enables robots to plan to complete tasks safely and assess risks when executing tasks.
%can teach robots to quantitatively assess risks.
% to train a logistic regression based \textbf{\sar \rmaum}.
% The novel \hrrt approach ensures the acting agents %humans and robots
% % to mitigate risks by challenging assumptions in each others' models and 
% % ensuring all agents on the team
% have representative models of the safety concerns in the domain.
% We focus on how robots can quantitatively assess risks.
The \HRRTing exercise guides the human-robot team to improve its mental model of the problem domain to characterize risks.
% A \HRRTed dataset is used to train a \rmaum, which allows the robot to predict the utility of actions to
% % choose appropriate actions to take in order to 
% mitigate risks during task execution.
% As part of the robot's \sar, we expect the robot to engage in dialogue about safety by providing clear and explainable reports about its risk assessment, reasoning, and behavior in order to earn the trust of human operators, as seen in \fref{fig:teaser}.
We test the robot's ability to symbolically plan tasks safely in several domains and to assess risks
% of our \rmaum 
in two domains---lunar habitat and household---with varying definitions of safety,
% problem domains under different risk conditions, 
and find that the \HRRTing paradigm effectively teaches the robot to accurately assess and mitigate risks.
% model accurately assesses and mitigates risks and the robot provides understandable reports explaining the evaluated risks.
Our work demonstrates the feasibility of \HRRTing for \sar in different domains, furthering robots' capabilities of acting on human-robot teams in collaborative tasks.
% demonstrates the power of deploying a simple explainable \rmaum in different domains and furthers robots' capabilities of acting in collaborative tasks.
% note that problem formulation should not cover \textit{all} situations or ensure actions are \textit{always} safe or make any \textit{guarantees} about safety, but should allow robot to reason over safety, quantify/evaluate safety, and communicate about safety

% \todo{make contributions clear}
% \begin{itemize}
%     \item \textbf{HRRT}
%     \item RMAUM probably not a contribution?
%     \item NL instructions may just confuse the point; may be best to pull this out to avoid confusing the reader from the main point
% \end{itemize}

% \todo{teaser figure of diff robots in diff environments identifying/assessing risks and reporting it}

\section{RELATED WORK}
\label{sec:related_work}

\subsection{Safety in Robotics Applications
% Safety and Language in Robotics Applications
}
\label{subsec:related_work_safety}

A large body of robotics research has greatly improved robots' capabilities of performing different tasks.
However, the safety of the executed tasks are often not considered~\cite{Zhang2020}.
% truly lend themselves to the deep understanding of the tasks to which they are being applied~\cite{Storks2021}.
% Furthermore, in safety-critical applications, the potential for anything and everything non-expert users say to trigger a robot action does not necessarily result in a safe and user-friendly assistive robot.
% Some works that apply large language models even acknowledge that safety of the commanded movements
% and the effect of the agent's action on user experience
% are not currently addressed in their work~\cite{Zhang2022}.
This indicates that much state-of-the-art robotics research
% current work on natural language in robotics applications
is limited in its use in safety-critical domains.

% motivate need for safety
Though safety has not been addressed,
% in robotic control through language,
research emphasizes the importance of safety when robots work alongside humans~\cite{Liu2020, Lim2000, Bozhinoski2019, Chen2022, Bogue2017, Dhillon1993, Dhillon2002, Guiochet2017, Hentout2019, Lasota2017, Schulte2020, Zacharaki2020}, especially in collaborative and assistive tasks~\cite{Billard2019, Bogue2017}.
% There is clearly a need for robot systems that can reason over language in a safety-aware manner.
Robot systems need to perform tasks safely.
% reason over language in a safety-aware manner.
In this work, we take inspiration from government and industry safety cultures and risk assessment standards---such as Failure Mode Effects Analysis (FMEA) \cite{FMEA, NASAFMEA} % FMEA_software, FMEA_UT, NASAFMECA
and root cause analysis \cite{RCA, OSHA_RCA}---to provide a principled way for robots to reason over safety~\cite{NASASafety, NASARiskManagement}. % NASASafetyWebsite, NASARiskManagementWebsite
We expect robots to understand how risks can be assessed according to the likelihood and consequences of undesirable events~\cite{NASAJPR, Guevara2024} % Peace2017
and to enact appropriate risk reduction strategies where possible~\cite{NASANPR}.

\subsection{Trust and Cooperation}
\label{subsec:related_work_trust_cooperation}

% role of trust and cooperation
We aim to allow humans and robots to work together to accomplish tasks safely.
Robots %and automation acting
in collaborative tasks are often not relied on appropriately, specifically when reliance on the system does not match the robot's true capabilities~\cite{Lee2004}.
% engage in a dialogue about safety in collaborative tasks.
\textit{Cooperative} tasks---in which agents work together to achieve positive-sum ``win-win'' outcomes---involve vulnerability~\cite{Kuipers2023}.  The social nature of cooperative tasks~\cite{Lee2004, Robinette2016} makes \textit{trust} between robot and human user crucial~\cite{Kuipers2022}.
% Robots %and automation acting
% in cooperative tasks are often not relied on appropriately, specifically when reliance on the system does not match the robot's true capabilities~\cite{Lee2004}.
% overtrust/distrust
We expect robots to participate in the ``social exchange relationship'' associated with interpersonal trust~\cite{Lee2004}. %, discussed in more detail in the following sections.
For robots to be trusted in cooperative tasks, they must 
% demonstrate that they are trustworthy and
earn the trust of their fellow agents~\cite{Kuipers2022}.
Inexplicable robot actions
% Robots taking actions that cannot be meaningfully explained 
will cause user distrust %users to distrust the robot 
\cite{Hopko2024, Lee2004, Robinette2016} as unreliable robots can also be unsafe~\cite{Dhillon1993}.
Eroded trust
% and a reputation for untrustworthiness
leads to robot disuse
% makes a robot more prone to disuse
\cite{Lee2004}
% and less likely to be used in cooperative tasks in the future 
in future cooperative tasks~\cite{Kuipers2022}.
% To %be safe and
% earn other agents' trust, robots must clearly explain their actions. % provide clear explanations for any action it takes.
% Researchers often aim to diagnose when failures may occur in robot systems~\cite{Zhang2020} and identify multiple types of failures or exceptions~\cite{She2015}.
% Inspired by work exploring multiple levels of representation~\cite{Kuipers2000, Beeson2010} and bridging the gap between these levels~\cite{Philippsen2009, Ziegler2015}, we expect robots to explain decisions and assessments of unsafe circumstances~\cite{Zhang2020, She2015} in terms of high-level causal details and low-level control details to ensure that the robot's \sa decisions are explainable regardless of the user's level of expertise.
We expect robots to report safety assessments
% of unsafe circumstances
\cite{Zhang2020, She2015} to human operators
% in terms of high-level causal details and low-level control details~\cite{Kuipers2000, Beeson2010, Philippsen2009, Ziegler2015} 
to ensure that robots' \sa decisions are explainable. % regardless of user expertise.

\subsection{Red Teaming}
\label{subsec:related_work_red_teaming}

% motivation for red teaming
Every agent in a cooperative task uses models to simplify the unboundedly complex world.
Simplifying models are necessary, but incomplete knowledge carries risk and ``unknown unknowns'' can cause disastrous outcomes~\cite{Kuipers2022}.
% While simplifying models are necessary, the incomplete knowledge of these models carries risks, and disastrous outcomes occur when some ``unknown unknown'' arises that is unaccounted for in the model~\cite{Kuipers2022}.
% Applied to \sar,
We want to minimize risks in the robot's incomplete knowledge to avoid unsafe situations and dangerous consequences.
% We propose that \sar requires \HRRTing to share autonomy between human and robot agents and challenge the assumptions made by both humans and robots in safety-critical domains.

% definitions and background for red teaming
\textit{Red teaming} detects weaknesses and vulnerabilities, explores possibilities, considers multiple perspectives, examines alternate analyses, reveals biases, and challenges conventional wisdom with adversarial perspectives~\cite{Kraemer2004, Longbine2008, Abbass2011, Schneier2012, Tan2014, Schneier2015, Zenko2015, Ganguli2022}.
% In red team exercises, 
The Blue Team (``good guys'')
% has an objective and
considers how the Red Team (``bad guys'') may thwart their objective, improving their approach
% .  The Blue Team can improves their plan or capabilities accordingly
to prevent %the Red Team's
attacks~\cite{Yang2006}.
Identifying ``upstream decision points''~\cite{Kuipers2022} can avoid dire consequences 
% through \textit{counter-factual reasoning},
% % Counter-factual reasoning is
% a form of causal inference~\cite{Pearl1995causaldiagrams, Pearl2009causality, Pearl2012docalculus} that considers 
by considering alternate versions of past events, and creating ``blueprints for future action''~\cite{Roese2017} % ~\cite{Roese1997}
to reach unrealized goals~\cite{Roese2011}. % Roese2008,
% , and explains complex systems~\cite{Wachter2017}.
Red teams improve decision making and mitigate risks~\cite{Longbine2008} before disastrous outcomes occur~\cite{Yang2006}.
Many domains use red teams,
% Red teaming has been applied in a number of domains,
including military~\cite{Longbine2008, Yang2006}, computer and cyber-security~\cite{Wood2000, Kraemer2004, Schneier2012, Adkins2013, Schneier2015, Mansfield2018}, and organizational procedures to challenge institutional biases~\cite{Zenko2015}.

% counter-factual reasoning
% Another option for
% % addressing vulnerabilities in a simplified model and exploring the vast space of possibilities~\cite{Yang2006, Abbass2011} is to
% avoiding dire consequences is identifying
% ``Upstream decision points''~\cite{Kuipers2022} can avoid dire consequences through \textit{counter-factual reasoning},
% % Counter-factual reasoning is
% a form of causal inference~\cite{Pearl1995causaldiagrams, Pearl2009causality, Pearl2012docalculus} for alternative versions of past events~\cite{Roese1997}, creating ``blueprints for future action''~\cite{Roese2017}
% % from alternate ways
% to reach unrealized goals~\cite{Roese2008, Roese2011}.
% Complex systems may use counter-factual explanations~\cite{Wachter2017}.
% Research has shown that counter-factual explanations effectively explain outputs of complex systems~\cite{Wachter2017}.
% We take inspiration from work that suggests explanations are crucial to building trust and counter-factual explanations are effective in complex systems~\cite{Wachter2017}.  We expect \sar to enable robots to consider alternate risk mitigating actions that could have been taken during a task, thereby identifying upstream decision points to avoid unsafe operating conditions.

% different types of red teams
Red teaming implementations vary with context, but focus on \textit{human} red teams, where humans simulate opponent viewpoints
% to challenge thinking
\cite{Wood2000, Kraemer2004, Longbine2008, Adkins2013, Zenko2015, Mansfield2018}.
% One work uses a human red team to generate adversarial examples and test the capabilities of a computational language model~\cite{Ganguli2022}.
More recent work explores
% \textit{automated} or
\textit{computational} red teams to automate creation of adversarial perspectives. % in different ways.
For example,
human~\cite{Ganguli2022} or computational~\cite{Perez2022} red teams can generate adversarial examples to evaluate computational models.
% computational models may be evaluated with adversarial examples generated by human red teams~\cite{Ganguli2022} or computational red teams~\cite{Perez2022}.
% For example, Perez \textit{et al.}~\cite{Perez2022} use a red language model to identify offensive language in a target language model.
% Abbass \textit{et al.}~\cite{Abbass2011} present multiple levels on which computational red teams may function and emphasize the importance of defining and modeling computational red teaming problems as \textit{multi-agent systems} that aim to explore the impact of each agent's actions on the system.
% Yang \textit{et al.}~\cite{Yang2006} use evolutionary algorithms in simulation to learn what blue team battlefield strategies minimize blue team harm against different fixed red team characteristics.
% The purpose of 
Computational red teams inform and focus human decision making
% in different domains
\cite{Yang2006, Abbass2011}, for example about physical security assessment of buildings~\cite{Tan2014} or defending against attacks exploiting vulnerabilities in large enterprise networks~\cite{Randhawa2018}.
Abbass \textit{et al.} \cite{Abbass2011} define levels on which computational red teams (CRTs) function:
\begin{enumerate*}[label=(\alph*)]
    \item \textbf{CRT0}: An agent equipped with a generic decision-making model does not evolve.
    \item \textbf{CRT1}: Each individual agent learns, adapts, and changes its decision-making process through interactions with the environment.
    \item \textbf{CRT2}: A team of agents learns and evolves together to defend against the fixed strategy of the opposing team.
    \item \textbf{CRT3}: Teams of agents evolve alongside an evolving environment.
    % The environment itself can change and evolve.
    \item \textbf{CRT4}: Agents and teams reflect to identify and unlearn their own biases.
\end{enumerate*}
\noindent These CRT levels inspire similar levels of analysis for our \hrrt paradigm, described in \sref{subsec:HRRT_paradigm}.

% human-robot red team
% Red teaming in complex systems is often less about making decisions and more about exploring the vast space of possibilities~\cite{Yang2006, Abbass2011}.
We take inspiration from red teams as ``reality checks'' throughout all stages of a procedure~\cite{Wood2000}.
Previous works focus on human red teams, computational red teams, and human teams informed by computational red teams.
For our work in \sar, computational agents alone should not make evaluative ethical or moral judgments \cite{Lee2004, Sheridan2016, Kuipers2018, Kuipers2020} that may affect human safely.
We propose a \hrrt paradigm in which humans and robots work together to challenge assumptions %provide checks-and-balances
in shared autonomy tasks.
% In particular, the \hrrt will challenge assumptions in both the humans' and robots' simplifying models, explore the possible threats to safety that may arise in a problem domain, select appropriate actions for mitigating risks, and identify upstream decision points to learn to avoid future unsafe situations.

\section{SAFETY-AWARE REASONING \\ PROBLEM FORMULATION}
\label{sec:problem_formulation}
% computational model

% \todo{be clear about terminology}
% \begin{itemize}
%     \item hazard (state) vs risk (quantitative measure)
%     \item risk condition is a state
%     \item consequence is an event that can occur with some probability in a given state
%     \item consider a chain of events; mowers leave gate open (risk) the dog got out (consequence) and went out on the ice on the river (risk) where it fell in and drowned (consequence)
% \end{itemize}

% \todo{make this a more general formulation about safety; the \rmaum is just our implementation of risk assessment, but the focus is on HRRT as a whole}
To reason over safe task execution, robots must understand hazards in the problem domain and risk mitigating actions to minimize the risks and progress towards task completion.
We present \sar, which includes the following components:
\begin{enumerate*}[label=(\alph*)]
    \item hazard identification,
    \item risk assessment,
    \item risk mitigation, and
    \item safety reporting.
\end{enumerate*}
The robot must consistently perform these sub-tasks in order to operate safely.
These components of \sar are informed by the \hrrt, which allows human-robot teams to explore the space of possibilities in safety-critical problem domains.

\section{METHODS}
\label{sec:methods}

\subsection{Human-Robot Red Teaming Paradigm}
\label{subsec:HRRT_paradigm}

\begin{figure*}
    \centering
    \includegraphics[width=0.70\textwidth]{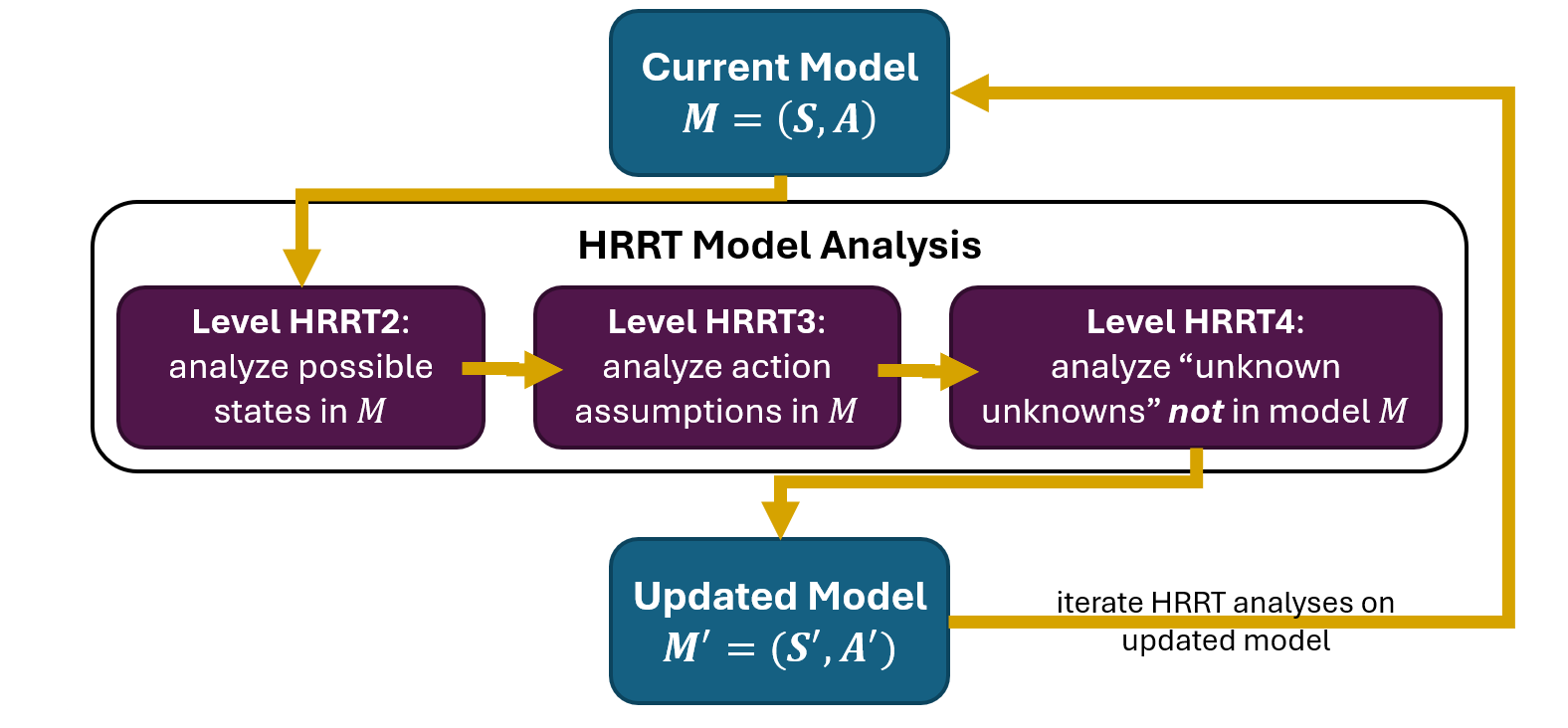}
    % {images/HRRT-outline-figure-v1.PNG}
    \caption{Overview of the \HRRT levels within the \HRRTing paradigm.  Each level analyzes different components of the modeled and unmodeled knowledge, creating an updated model that can be further iterated upon.}
    \label{fig:hrrt-levels}
\end{figure*}

We present the \HRRTing paradigm.
The human-robot team iterates over models of the environment, composed of a set of symbolic states $S$ and actions $A$.  We define a model $M$ as:
\begin{equation}
    \label{eq:hrrt_model}
    M=(S,A)
\end{equation}
\noindent based on the sets of states and actions that describe the robot's reasoning in that environment.  The model $M$ may take many forms---for example, a Markov Decision Process (MDP) or Partially Observable Markov Decision Process (POMDP)~\cite{Littman2013, RussellNorvig2020}. % lane2001toward, RussellNorvig2010
% In fact, as one example of the application of the updated model, we train a logistic regression model based on the data generated by the \hrrt to predict the utility of performing risk mitigating actions, as described in \sref{subsec:HRRT_robot_exps}.
We make no assumptions about the form of the model. The purpose of the \HRRT is not to work within a particular model formulation, but to explore what may be excluded from the model that will prove to be critical in a task that requires safety. If we consider the space of all possible models $\mathcal{M}$, we want to determine what models $M' \in \mathcal{M}$ will provide more information than current model $M$ about safe task performance in a given problem domain.

% The purpose of the mental model $M$ is to allow the robot to reason over states and actions while making a closed-world assumption, meaning that anything not included in the model is assumed to be negligible to the task.  However, even if the model itself represents a closed world, it does not necessarily include all knowledge an agent or team of agents may have.  Therefore, the \hrrt will also maintain a knowledge base $K$ in the space of all knowledge in the universe $\mathcal{K}$ of information relevant to safely achieving task goals.  For example, organizational guidelines or industry standards of safety may be relevant for the knowledge base.

We define levels of \HRRTing, similar to the levels describing computational red teams \cite{Abbass2011} (\sref{subsec:related_work_red_teaming}).
% , \HRRTing will similarly benefit from multiple levels of abstraction, similar to those used to describe computational red teams \cite{Abbass2011} (summarized in \sref{subsec:related_work_red_teaming}).
% As described in \sref{subsec:related_work_red_teaming}, we propose that \HRRTing The capabilities of computational red teams are described according to several levels of reasoning \cite{Abbass2011}. We suggest that \HRRTing will similarly benefit from multiple levels of abstraction to categorize future research and describe the responsibilities and capabilities of different human-robot teams.
\textit{Human-robot red teams} are specific subsets of CRTs, where computational agents work on teams alongside humans.
We observe that for computational red teams, levels CRT2, CRT3, and CRT4 describe \textit{teams} of agents.
Taking inspiration from these levels, we propose 3 similar stages for \hrrts (\HRRTs), summarized in \fref{fig:hrrt-levels} and described in the following sections.
% The following sections describe the characteristics of \hrrts at each of the proposed levels, summarized in \fref{fig:hrrt-levels}.

\subsubsection{Human-Robot Red Teaming Level 2}
\label{subsec:HRRT_level2}

A \hrrt operating on level HRRT2 explores possible scenarios and outcomes according to the team's shared knowledge of the environment.
% Red robot agents on level HRRT2 help the team enumerate possibilities that may have previously been unconsidered.
% based on their given knowledge of the environment.
% By enumerating the possible risky scenarios that could be encountered during task execution, the robot agent prompts the human-robot blue team to consider appropriate responses.  As a result, the human-robot team learns appropriate strategies for acting safely in the environment.
In particular, at level HRRT2, the \hrrt reasons over the current model $M=(S,A)$ and generates a list of possibilities:
\begin{equation}
    \label{eq:hrrt2}
    \mathcal{H}_2: \mathcal{M} \rightarrow S\times A \times S
    % \{(s,a,s')\}
\end{equation}
\noindent where each possibility is a tuple $(s,a,s') \in S \times A \times S$ representing possible state action transitions supported by the model. The function $\mathcal{H}_2$ algorithmically generates these possibilities by considering each possible symbolic state $s \in S$,
% $s \in \mathcal{P}(S)$ where $\mathcal{P}(S)$ is the powerset of all possible states, 
any action $a \in A$ that can be taken in each of those states, and the next state $s' \in S$ induced by taking action $a$ in state $s$.
Based on these possibilities, the human-robot blue team may update the model to reinforce valid possibilities, prevent invalid possibilities, or consider appropriate responses to unlikely events, creating an updated model $M_2'$.

\subsubsection{Human-Robot Red Teaming Level 3}
\label{subsec:HRRT_level3}

A \hrrt operating on level HRRT3 further analyzes its knowledge of the environment by challenging assumptions made by the team's model of the world.
% Modeled knowledge such as state information and definitions of actions implicitly include assumptions about the world.
% By challenging these assumptions, red robot agents prompt the human-robot blue team to modify or loosen these assumptions and consider contingency plans.  As a result, the human-robot team improves its ability to plan around unexpected circumstances.
At level HRRT3, the \hrrt identifies implicit assumptions in the team's model $M=(S,A)$, specifically whether pre-conditions for an action $a\in A$ can be reached to perform the action and whether post-conditions for action $a$ are expected to be achieved as a result of performing the action. %In particular, 
The function $\mathcal{H}_3$ generates
% maps the current model to 
pre- and post-condition assumptions:
\begin{equation}
    \label{eq:hrrt3}
    \mathcal{H}_3: \mathcal{M} \rightarrow \Omega_\textrm{pre} \times \Omega_\textrm{post}
\end{equation}
\noindent where $\Omega_\textrm{pre}=\{\omega_\textrm{pre}\}$ is a set of pre-condition assumptions $\omega_\textrm{pre} = (\{s\}, a)$ and $\Omega_\textrm{post}=\{\omega_\textrm{post}\}$ is a set of post-condition assumptions $\omega_\textrm{post} = (a, \{s\})$ implied by a given action.
% A pre-condition assumption $\omega_\textrm{pre} \in \Omega_\textrm{pre}$ has the form:
% \begin{equation}
%     \omega_\textrm{pre} = (\{s\},a)
% \end{equation}
% \noindent where $\{s\}$ is the set of pre-condition states required to perform action $a$.
% A post-condition assumption $\omega_\textrm{post} \in \Omega_\textrm{post}$ has the form:
% \begin{equation}
%     \omega_\textrm{post} = (a, \{s\})
% \end{equation}
% \noindent where $\{s\}$ is the set of post-condition states likely to result from performing action $a$.
For these assumptions $\omega \in \Omega$, the ordered pair of states and actions indicates an assumed causal link within an action.
These assumptions can be algorithmically generated by the red robot agent based on each pre- and post-condition for an action $a$.
Based on these assumptions, the human-robot blue team may update the model to modify actions, add additional validation actions, or add information for contingency planning around unsatisfied conditions.
These updates result in a modified model $M_3'$, improving its ability to plan around unexpected circumstances.

\subsubsection{Human-Robot Red Teaming Level 4}
\label{subsec:HRRT_level4}

A \hrrt operating on level HRRT4 learns from the previous analyses
% performed on levels HRRT2 and HRRT3
and improves its modeled knowledge as a result.
% The goal of level HRRT3 is to challenge assumptions and create contingency plans for when assumptions do not hold.
\iffalse
The goal of level HRRT4 is to recognize that possibilities on level HRRT2 and assumptions on level HRRT3 may reveal umodeled factors that will affect task completion, and improve the modeled knowledge accordingly.
Models necessarily simplify the infinite complexity of the real-world.
But these simplifications or ``unknown unknowns'' become catastrophic when something left out of the model proves to be important~\cite{Kuipers2022}.
By improving the team's model and accounting for new knowledge, the human-robot team improves its ability to reason and complete tasks in a safety-critical environment.
Level HRRT4 takes information about possibilities and assumptions from levels HRRT2 and HRRT3 to create an updated model $M'$.
\fi
In particular, the $\mathcal{H}_4$ function takes in the current model $M$, the enumerated possibilities from $\mathcal{H}_2$, the assumptions from level $\mathcal{H}_3$, and a dialogue tree $\Sigma$:
\begin{equation}
    \label{eq:hrrt4}
    \mathcal{H}_4: \mathcal{M} \times \mathcal{H}_2(M)\times\mathcal{H}_3(M) \times \Sigma \rightarrow \mathcal{M} %\times \mathcal{K}
\end{equation}
\noindent where the dialogue tree $\Sigma$ (inspired by \cite{DOROTHIE2022})
% (described in \sref{sec:methods_dialogue}) 
prompts deeper reflections for the human-robot team.
In our implementation, the dialogue $\Sigma$ is implemented as a simple English-like interface that allows the \hrrt to ask the human-robot blue team probing questions, such as
% . The interactions in $\Sigma$ may be 
general safety questions (for example, ``Are there external, independently verified resources for identifying failure cases in this domain?'') or more domain-specific questions.
Using the interactions in $\Sigma$, the \hrrt takes the current model $M\in \mathcal{M}$ and previous analyses
% the supported possibilities 
$\mathcal{H}_2(M)$ and 
% implied assumptions
$\mathcal{H}_3(M)$ to ask the human-robot blue team for insight on weaknesses and limitations in the model,
% . The interactions in $\Sigma$ will further 
prompting the team to create
% by adding new states or actions or modifying existing states or actions to address assumptions and undesired possibilities.
% This creates an 
updated model $M_4' \in \mathcal{M}$. % as well as additional knowledge for the knowledge base $K' \in \mathcal{K}$.

% The analysis performed by the \HRRT on level HRRT4 is meant to be a reflection on the insights from the previous levels as well as an opportunity to address identified shortcomings of the model.
% As described in \sref{sec:CRT_limitations}, 
This HRRT4 reflective process cannot be completely automated due to the limitations of computational teams, namely
% in completing these high-level reflective exercises
the need for humans in the decision-making loop to make ethical or moral judgments \cite{Lee2004, Sheridan2016, Kuipers2018, Kuipers2020}.
The human insight in the \HRRT process is necessary to help direct and prioritize improvements while generating updated models $M'$.

\subsubsection{Iterating through \HRRT Levels}
\label{subsec:HRRT_iterations}

Since the \HRRTing levels allow the team to adapt its modeled knowledge, we expect these levels to iteratively repeat.
More specifically, an iteration is one HRRT2 analysis, one HRRT3 analysis, and one HRRT4 analysis, as depicted in \fref{fig:hrrt-levels}.
% Repeated iterations means proceeding through all 3 levels multiple times.
% Given the updated domain knowledge achieved on level HRRT4, the team will return to level HRRT2 in order to enumerate possibilities again based on the newly acquired information.
% This enumeration of possibilities will teach the \hrrt to operate safely given its new knowledge.
% Upon returning to level HRRT3, the team may realize its improved domain knowledge contains additional limiting assumptions that must be challenged.  The team can consider alternate perspectives and create contingency plans.
These repeated analyses could highlight errors in the model, identify additional ``unknown unknowns'' that must be accounted for, and consider the long-term implications of modeled knowledge in order to improve the human-robot team's ability to perform tasks safely.
% Iterating through the levels enables the team to not only consider unaccounted for factors, but also consider the long-term implications of new knowledge added to the model.
Every time the \hrrt modifies its modeled knowledge and reflects on unmodeled factors, the \HRRT analysis should be repeated.
% , the possibilities accounted for by that model must be enumerated and assumptions implicit in that knowledge should be challenged accordingly.

% We chose to define one iteration as one analysis on each \HRRT level because the analyses are often related.
% For example, repeating several analyses on level HRRT2, then repeating several analyses on level HRRT3 as opposed to proceeding straight from level HRRT2 to HRRT3 may limit the scope of analysis to that level's perspective rather than the complete knowledge of the model. We further investigate the value of each individual level and iterations through all levels in \sref{subsec:ablation_study}.

% On level HRRT4, the function $\mathcal{H}_4$ generates an updated model $M_4'$. If we repeat the \HRRTing process, 
Each iteration through the levels produces a \textit{model hypothesis}, which contains more information than the previous model.
We define a model hypothesis $M^i$ generated by iteration $i$ through the \HRRT levels.
% The purpose of simplifying models is to simplify reasoning in an infinitely complex world. As a result, it is 
We may prefer to use a simpler model $M^i$ generated at iteration $i$ that solves the same problem as a more complex model $M^j$ generated at iteration $j$ where $i < j$. However, we do not want the human-robot team to disregard more complex models $M^j$ since these models may include edge cases or remote possibilities that prove to be critical in high-risk problem domains.
Therefore, the \hrrt takes a hybrid approach to these mental models. In particular, the \hrrt maintains a set of generated model hypotheses $\{M^i\}_{i=0}^N$ for each of the $N$ iterations through the \HRRT levels. Each model may be useful for solving different types of problems in the problem domain. We evaluate the value of a hybrid model in \sref{subsec:planning_exps}.

% removed example
\iffalse
For example, the robot trying to get through the locked habitat door without its keycard now has the option to pound down the door.  Its model now includes the immediate consequences of this action, namely that the external habitat door is broken and the airlock may be damaged.  While iterating through the \HRRT levels, the \hrrt will realize that there are additional consequences such as sudden unexpected depressurization of the airlock.  The \hrrt will also realize that there are longer-term consequences, namely that the crew no longer has a way to safely get outside without depressurizing the whole habitat.
There were unforeseen long-term consequences to this new action of pounding down the door that would not have been caught without iterating back through the \HRRT levels.  With these iterations, the \hrrt can revisit contingency plans (for example, what if a prerequisite for leaving the habitat in the first place is to have the keycard), update the modeled knowledge, and explore the long-term consequences of this new information.
\fi

% Since the world is infinitely complex, iterating over the levels of \HRRTing could become an intractable problem.
% However, all models make simplifications based on what can be assumed to be negligible.
% Similarly, 
The iterations through the \HRRTing levels will terminate when the factors being considered for inclusion within the model are negligible %.
% additional model knowledge is negligible.
% When there are no new possibilities to enumerate on level HRRT2, when the team determines that all assumptions are valid at level HRRT3, and when no new knowledge is added to the model on level HRRT4, then the \hrrt can conclude its model is 
% adequate %sufficiently complex
% for the problem domain.
% Anything not included in a computational model of the world is assumed to be negligible.  This is not to say that the symbols or actions not in the model do not exist, just that they are not particularly important to the intended task.
% For example, the robot trying to get through the lunar habitat door likely does not need to account for 
% someone spilling a hot coffee on the floor in mission control. %an asteroid crashing through the ceiling.
% It is not that the event is impossible, just that such an event can be disregarded from the task the robot is trying to complete.
% % (for example, because the asteroid would likely destroy the door anyway).
based on the discretion of the human-robot team.
% Factors that can be considered negligible are up to the discretion of the human-robot team and will vary based on the problem domain and environment.
Iterating through the \HRRTing levels does not itself solve the problem of ``unknown unknowns'' in computational models.
However, the \HRRT levels provide additional opportunities to explore possibilities, challenge assumptions, and update domain knowledge, which is especially important in extreme environments and safety-critical problem domains.

\subsection{Composition of Human-Robot Red and Blue Teams}
\label{subsec:methods_team_composition}
% We describe the composition of our human-robot red and blue teams in the following subsections.
The \hrrt will probe the human-robot blue team to expand its knowledge of different problem domains.
The focus of our work is on the methods that the \hrrt uses to query and prompt the human-robot blue team.
In our implementation, the red computational agent is a chatbot that uses a dialogue tree $\Sigma$ for simple English-like interactions.
To avoid biasing our results in favor of the red team, we minimize red human inputs and rely on the red computational chatbot agent to challenge the human-robot blue team's understanding. % of the problem domain.

% To simulate the human-robot blue team, w
We used ChatGPT \cite{ChatGPT} as the blue computational agent.
ChatGPT is the state-of-the-art in computational agents engaging in natural language question-answer interactions.
% Due to ChatGPT's advanced interactions, a
All new symbols (states and actions) presented into the model were generated by ChatGPT %the computational agents
% All modifications to the models were proposed by ChatGPT 
as a direct result of the prompts from the red computational chatbot agent.
ChatGPT would make many suggestions, but often struggled to make actionable changes to the model. % all of the suggestions at once. %, as described in \sref{sec:CRT_limitations}.
To assist, the blue human agent (one of the authors) would focus the blue computational agent to 2-5 suggested modifications at each level.

\section{EXPERIMENTS AND RESULTS}
\label{sec:experiments_results}

\subsection{\textit{Safety-Aware Reasoning} Symbolic Planning Experiments}
\label{subsec:HRRT_plan_exps}

We first evaluate how the \HRRTing approach can help robots plan safe tasks. % to perform tasks safely.
% We selected several problem domains, each with varying definitions of safety and risks.
% Our selected domains will be described in \sref{subsec:planning_domains}.
% \sref{subsec:exps_iterations} describes the iterations through the \HRRT levels.
% We describe our ablation study over the \HRRT levels in \sref{subsec:ablation_study} and model saturation experiments in \sref{subsec:model_saturation}.
% The results of our symbolic planning experiments will be shown in \sref{subsec:planning_exps}.

\subsubsection{Safety-Critical Planning Domains}
\label{subsec:planning_domains}

To evaluate the value of the \HRRTing approach, we test our methods in a variety of problem domains.  We considered problem domains with varied levels of risk and different definitions of safety:
\begin{enumerate*}[label=(\alph*)]
    \item \textbf{Space: Lunar Habitat} (a robot assists astronauts living in a pressurized habitat to conduct science experiments on the lunar surface);
    % is on the lunar surface, assisting astronauts living in a pressurized habitat with science experiments and space exploration tasks. %The pressurized lunar habitat has an airlock, with two doors on either side. One door from the pressurized habitat to the airlock chamber, one door from the airlock chamber to the lunar surface.
    \item \textbf{Space: Mars Science Team} (a team of robots communicating with ground control on Earth conduct science experiments on Mars to learn about long-term presence in space); %The robot is working on a team of robots, and they have communication with ground control on Earth. Due to the long distance between Earth and Mars, the robots must handle time-delayed communications.
    \item \textbf{Household: Assembly and Repairs} (a robot performs regular home maintenance, assembly, and repair tasks); %including any assembly tasks or repairs, to assist the family that lives there. %For example, the robot may help assemble a new desk chair or fix a broken cabinet door. The robot needs to safely use tools to assess and complete these maintenance tasks.
    \item \textbf{Household: Cleaning} (a robot cleans a house within which a 
    % is going to clean every room in the house. A 
    family of humans, including curious children and pets, live); % in the house as well. %The robot needs to effectively complete the cleaning tasks while keeping everyone in the house safe and preventing children and pets from getting into any dangerous cleaning chemicals.
    \item \textbf{Everyday: International Travel} (a robot personal assistant helps a human plan a trip); %, which will require international travel. %The robot needs to plan the trip, adapt to unforeseen issues that may arise, and help the human prepare to travel safely.
    \item \textbf{Everyday: Vehicle Maintenance} (a robot personal assistant helps a human diagnose issues with their vehicle); %. %The human uses the vehicle to drive to-and-from work everyday as well as complete required errands. The robot needs to help the human diagnose and correct issues to keep the vehicle in good working order and safe for operations.
    \item \textbf{Cinematic: Nuclear Warfare} (a robot must protect human life from a nuclear missile attack, inspired by the movie \textit{The Iron Giant}~\cite{IronGiant}); %A robot is serving as an improved version of the wartime robot from the movie \textit{The Iron Giant}~\cite{IronGiant}. The robot is equipped with defensive weapons systems and flight systems and must protect human life from wartime risks such as nuclear missile attacks, resorting to self-sacrifice if needed.
    and
    \item \textbf{Cinematic: AI Captain} (a robot supports the success of space exploration mission objectives and protects human crew, inspired by the movie \textit{2001: A Space Odysssey}~\cite{SpaceOdyssey}). %; %---A robot is serving as an improved version of the HAL 9000 computer system from \textit{2001: A Space Odyssey}~\cite{SpaceOdyssey}. The improved robot system needs to support the success of human space exploration mission objectives without putting the lives of the human crew in jeopardy.
\end{enumerate*}
\noindent Collectively, these problems are meant to explore how our \HRRT approach performs in uncovering the complexities of different domains. % in order to plan to safely complete tasks.
% In each domain, the \hrrt challenged the human-robot blue team's thinking about these problems to uncover insights about how to perform tasks safely.
% Results for how the \HRRTing exercise improved the team's knowledge in these domains are presented in \sref{subsec:planning_exps}.

\subsubsection{\HRRT Iterations}
\label{subsec:exps_iterations}

For each of the 8 problem domains described in \sref{subsec:planning_domains}, we created a minimal starting model $M^0$ and 
% performed 5 iterations of
iterated through
the \HRRT levels (where an iteration is HRRT2, HRRT3, and HRRT4), generating updated models $\{M_4^i\}_{i=1}^N$. At each iteration $i$, our \HRRT implementation prompted the blue computational agent ChatGPT based on model possibilities $\mathcal{H}_2(M^i)$, model assumptions $\mathcal{H}_3(M^i)$, and queried model updates through $\mathcal{H}_4$ based on the English-like interactions $\Sigma$ from the red chatbot.
% \tref{tab:hrrt2_example}, \tref{tab:hrrt3_example}, and \tref{tab:hrrt4_example} illustrate example analyses that occur on each of the \HRRT levels (HRRT2, HRRT3, and HRRT4, respectively) within the Space Mars Science Team domain.
Our proposed \HRRT methods challenge the team's understanding of the domain and guide the team through iteratively improving the modeled knowledge.
% We give one example of a full iteration (\HRRT Level 2, Level 3, and Level 4) for each problem domain in \aref{app:HRRT_chatgpt}.
% \tref{tab:model_example} shows how the model for the Space Mars Science Team problem domain evolved over the 5 \HRRT iterations.
% The final \HRRTed models for each problem domain can be seen in \aref{app:HRRT_models}.
% The output from ChatGPT tends to be verbose and includes symbols from the low-level format of the model it was prompted to iterate on. As a result, we do not include the full logs of the \HRRT iterations here (though they are available in our code \ES{link to github}). However, to illustrate the types of interactions and reflections prompted by some of the questions posed by the red chatbot, we include the raw ChatGPT output for one iteration in each domain in \aref{app:HRRT_chatgpt}. \ES{ref to model appendix?}

\subsubsection{Ablation and Saturation Experiments}
% Study over \HRRT Levels
% }
\label{subsec:ablation_saturation_study}

We explored the value of the proposed \HRRT levels and iterations, focusing on the Mars Science Team.  After each level, we saved the generated model hypothesis.
We also investigated
% explore the value of the iterations through the \HRRT levels, specifically investigating 
whether successive iterations lead to saturation of the modeled knowledge.
We performed 10 full \HRRT iterations 
% (where an iteration is HRRT2, HRRT3, and HRRT4) 
for a total of 30 models across each iteration and level.
For both the ablation and saturation experiments, we tested how the models perform in 200 randomized planning tasks.

\begin{figure}
    \centering
    \includegraphics[width=0.48\textwidth]{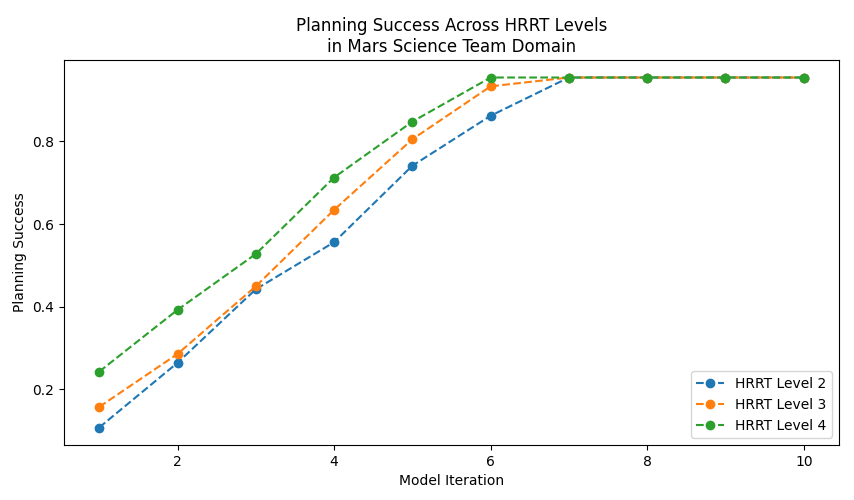}
    \caption{Results of the ablation study over the \HRRT levels and the saturation experiments over \HRRT iterations. Each ablation excludes the higher levels of analysis. We tested the models at each level and iteration in 200 randomized planning tasks in the Mars Science Team problem domain. We see that each \HRRT level builds on the knowledge gained from the previous levels.
    We also see that around \HRRT iteration 6, the modeled knowledge becomes saturated,
    % the human-robot teams started repeating many suggestions and analyses, 
    as reflected by the flattening of the curve. 
    % At this point, the modeled knowledge becomes saturated.
    % Successive iterations may provide insights on additional tasks, but do not fundamentally change the learned safety and risk mitigation mechanisms.
    }
    \label{fig:ablation}
\end{figure}

\fref{fig:ablation} summarizes the results of the ablation study over the \HRRT levels and the saturation experiments with successive iterations.
These results indicate that each level of analysis builds upon the previous levels, improving the modeled knowledge and the team's ability to handle planning problems in the domain.
This provides evidence to support our choice of defining an iteration as one HRRT2 analysis, one HRRT3 analysis, and one HRRT4 analysis. Since each level builds on each other, it is valuable to proceed through all levels of analysis, then iterate back through the levels to further analyze the modeled knowledge.

% To provide a more concrete illustration of how the \HRRT levels transform and evolve the modeled knowledge, we provide an example from this ablation study in \tref{tab:hrrt_iteration_example}.
% We see that every time updates are added to the model, the next iteration through the \HRRT levels identifies new relevant information, allowing the team to evolve its understanding of what it means to complete tasks in the problem domain.
% The interrelated analyses depicted in this example further justifies the definition of an \HRRT iteration as one HRRT2 analysis, then one HRRT3 analysis, and then one HRRT4 analysis.
% Transitioning between the levels uncovers the interrelated issues of possibilities, assumptions, and additional model knowledge, rather than getting stuck at just one level of analysis.
% Since each level is related and builds on the information from the previous level, it is important to consider each analysis together in one iteration.

% \subsubsection{Model Saturation Experiments}
% \label{subsec:model_saturation}

% We then explore the value of the iterations through the \HRRT levels, specifically investigating whether successive iterations will lead to saturation of the modeled knowledge.  As with the ablation study in \sref{subsec:ablation_study}, we focused on the Mars Science Team domain.  We completed 10 \HRRT iterations and tested the models in 200 randomized planning problems.

% \input{figures_tables/saturation_figure}

% \fref{fig:saturation} summarizes the findings of the saturation experiment. 
We also see that by \HRRT iteration 6, the model becomes saturated. The outputs from the \HRRTing exercise started to repeat at this point, and the model contained sufficient risk mitigation mechanisms to plan safely with a high success rate.
We expect the saturation point will vary with the complexity of the environment and knowledge of agents of the team. However, these experiments provide evidence to support that successive iterations through the \HRRT levels allow the human-robot team to gain sufficient insight to plan tasks safely in a given problem domain.

\begin{table}[ht]
    \centering
    \begin{tabular}{|c|c|c|c|c|}
        \hline
        \textbf{Domain} & \textbf{Problem} & \textbf{Planning} & \textbf{Total} & \textbf{Success} \\
        \textbf{Class} & \textbf{Domain} & \textbf{Successes} & \textbf{Tasks} & \textbf{Rate} \\
        \hline \hline
        \multirow{2}{*}{Space} & Lunar Habitat & 49 & 50 & 0.98 \\
         & Mars Science Team & 43 & 50 & 0.86 \\
        \hline
        \multirow{2}{*}{Household} & Assembly/Repairs & 50 & 50 & 1.00 \\
         & Cleaning & 44 & 50 & 0.88 \\
        \hline
        \multirow{2}{*}{Everyday} & International Travel & 46 & 50 & 0.92 \\
         & Vehicle Maintenance & 47 & 50 & 0.94 \\
        \hline
        \multirow{2}{*}{Cinematic} & Nuclear Warfare & 32 & 50 & 0.64 \\
         & AI Captain & 39 & 50 & 0.78 \\
        \hline \hline
        \multicolumn{2}{|c|}{\textbf{TOTAL}} & \textbf{350} & \textbf{400} & \textbf{0.875} \\
        \hline
    \end{tabular}
    \caption{Cumulative \sar planning experiments demonstrating the \HRRT approach.}
    \label{tab:planning_summary}
\end{table}

\subsubsection{Safety-Critical Planning Experiments}
\label{subsec:planning_exps}

For each problem domain, we performed 5 \HRRT iterations, generated model hypotheses $\{M_4^i\}_{i=0}^5$, %for each problem domain
% described in \sref{subsec:planning_domains} 
% were generated through the \HRRT iterations, we
% the human red agent (the researcher) 
and converted the \HRRTed models 
% blue team's (ChatGPT's) suggestions
into the Planning Domain Definition Language (PDDL) \cite{PDDL1998}.
We investigated failure cases for each domain through external independent documentation \cite{SafetySpace, SafetyNuclear, SafetyAI} % SafetyAI_IBM, SafetyHouseRepairs, SafetyHouseCleaning, SafetyTravel, SafetyVehicle
% Based on these failure cases, we 
and generated 50 planning tasks per domain, where each initial state included a randomly generated subset of failure cases.
The PDDL descriptions of the domains (based on the \HRRTed models) and the planning tasks were given to an % symbolic STRIPS planner 
off-the-shelf symbolic STRIPS task planner\footnote{Pyperplan STRIPS planning library: \url{https://github.com/aibasel/pyperplan}} \cite{Pyperplan}
to evaluate whether the models were 
sufficient to plan around safety-critical failures and achieve task goals.
% sufficiently complex to plan around the safety-critical failures while achieving task goals.

\begin{figure}
    \centering
    \includegraphics[width=0.98\linewidth]{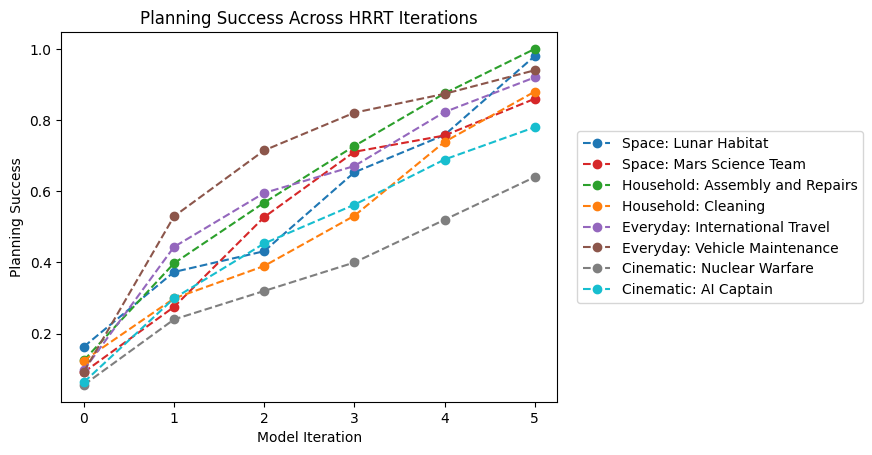}
    \caption{Planning problem success rates per iteration of the \HRRTed models across all domains.
    % This plot overlays the results from \fref{fig:model_iterations_space}, \fref{fig:model_iterations_household}, \fref{fig:model_iterations_everyday}, and \fref{fig:model_iterations_cinematic} for a comparative analysis between these domains.
    }
    \label{fig:cumulative_model_iterations}
\end{figure}

\tref{tab:planning_summary} summarizes the results for the planning tasks in each problem domain, aggregated over all of the generated model hypotheses.
% Figures depicting the contribution of each individual model hypothesis to the planning success in the domain are summarized for each problem class in \fref{fig:model_iterations_space} (space applications), \fref{fig:model_iterations_household} (household applications), \fref{fig:model_iterations_everyday} (everyday applications), and \fref{fig:model_iterations_cinematic} (cinematic applications).
\fref{fig:cumulative_model_iterations} depicts the impact of successive model iterations on planning success across all of the problem domains.
These results indicate the promise of our proposed methods for iterating through the levels of \HRRTing.
Each iteration through the levels made the generated model hypotheses more capable of handling failures. %, as seen in \fref{fig:cumulative_model_iterations}.
Our generated models successfully planned to achieve task goals, mitigate risks, and avoid critical failures with a success rate of 0.875 over a combined 400 planning tasks in 8 different problem domains (\tref{tab:planning_summary}).
% Additional \HRRT iterations and including blue human agents with insider or expert knowledge for the respective problem domains could further improve our results.
These results demonstrate that the proposed \HRRTing methods
% and the iterative \HRRT levels 
help human-robot teams %reflect on and
uncover the complexities of mitigating risks and avoiding ``unknown unknown'' failure cases in safety-critical problem domains.

\subsection{\textit{Safety-Aware Reasoning} Robot Execution Experiments}
\label{subsec:HRRT_robot_exps}

% \todo{edit this section down}

\begin{figure*}
    \centering
    \begin{subfigure}{0.3\textwidth}
        \centering
        \includegraphics[height=2.8cm,keepaspectratio]{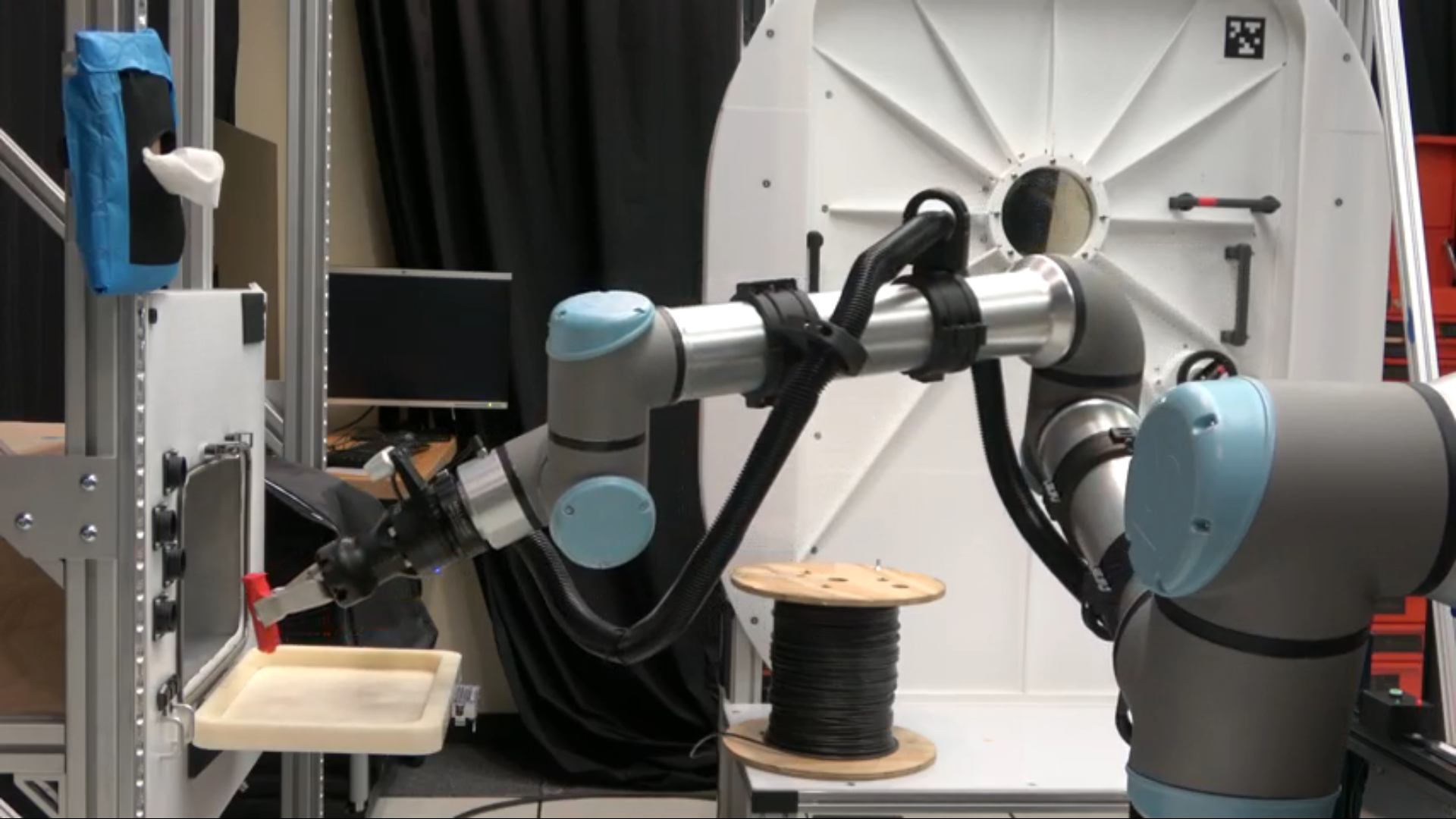}
        \caption{
        Robot safely performs the sample stowage task.
        % Safely performing the sample stowage task.
        }
    \end{subfigure}
    \hfill
    % \par\bigskip
    \begin{subfigure}{0.3\textwidth}
        \centering
        \includegraphics[height=2.8cm,keepaspectratio]{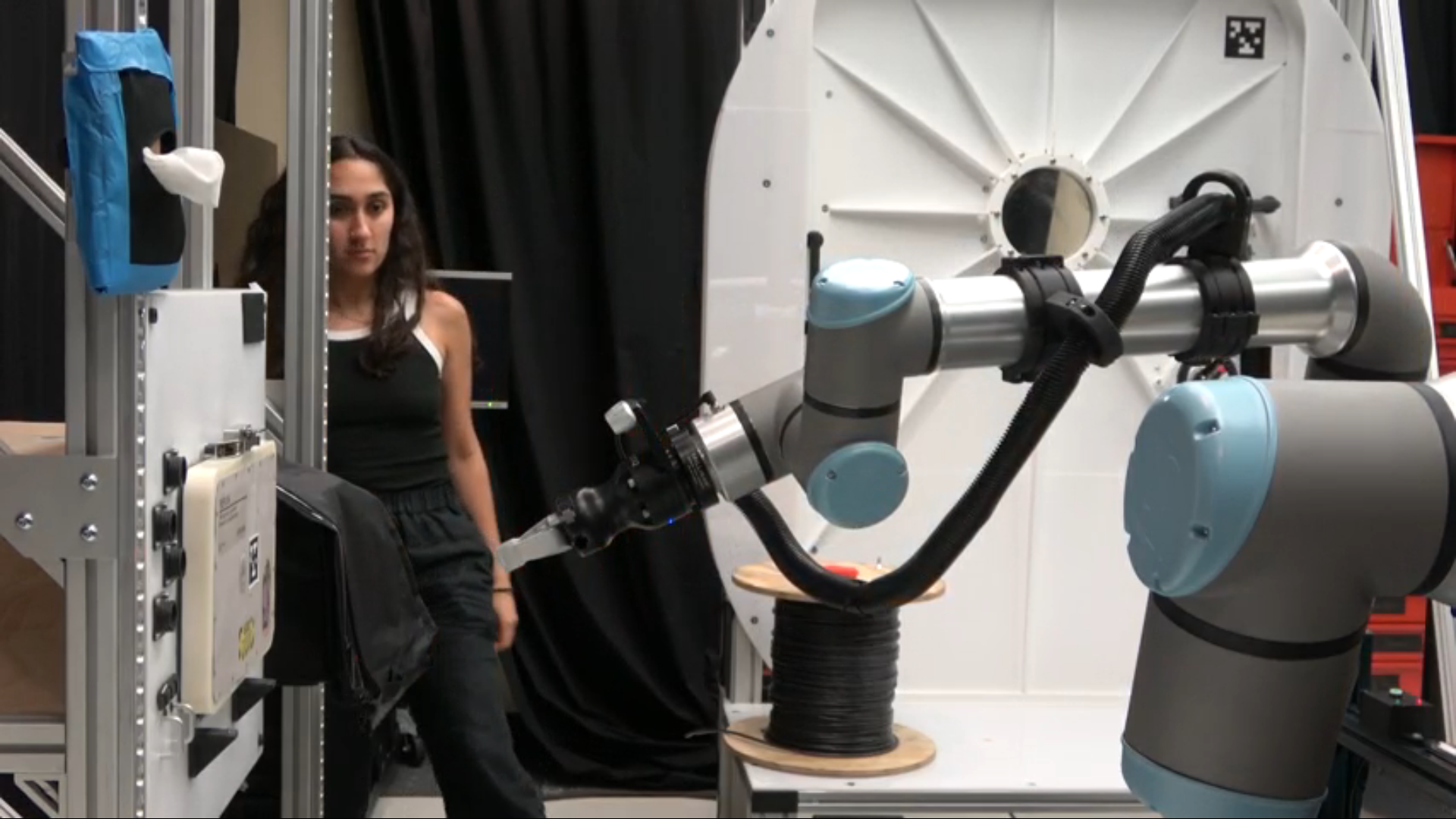}
        \caption{
        Robot aborts the task when a human astronaut enters the workspace.
        % The robot aborts the task because a human astronaut enters the workspace.
        }
    \end{subfigure}
    \hfill
    % \par\bigskip
    \begin{subfigure}{0.3\textwidth}
        \centering
        \includegraphics[height=2.8cm,keepaspectratio]{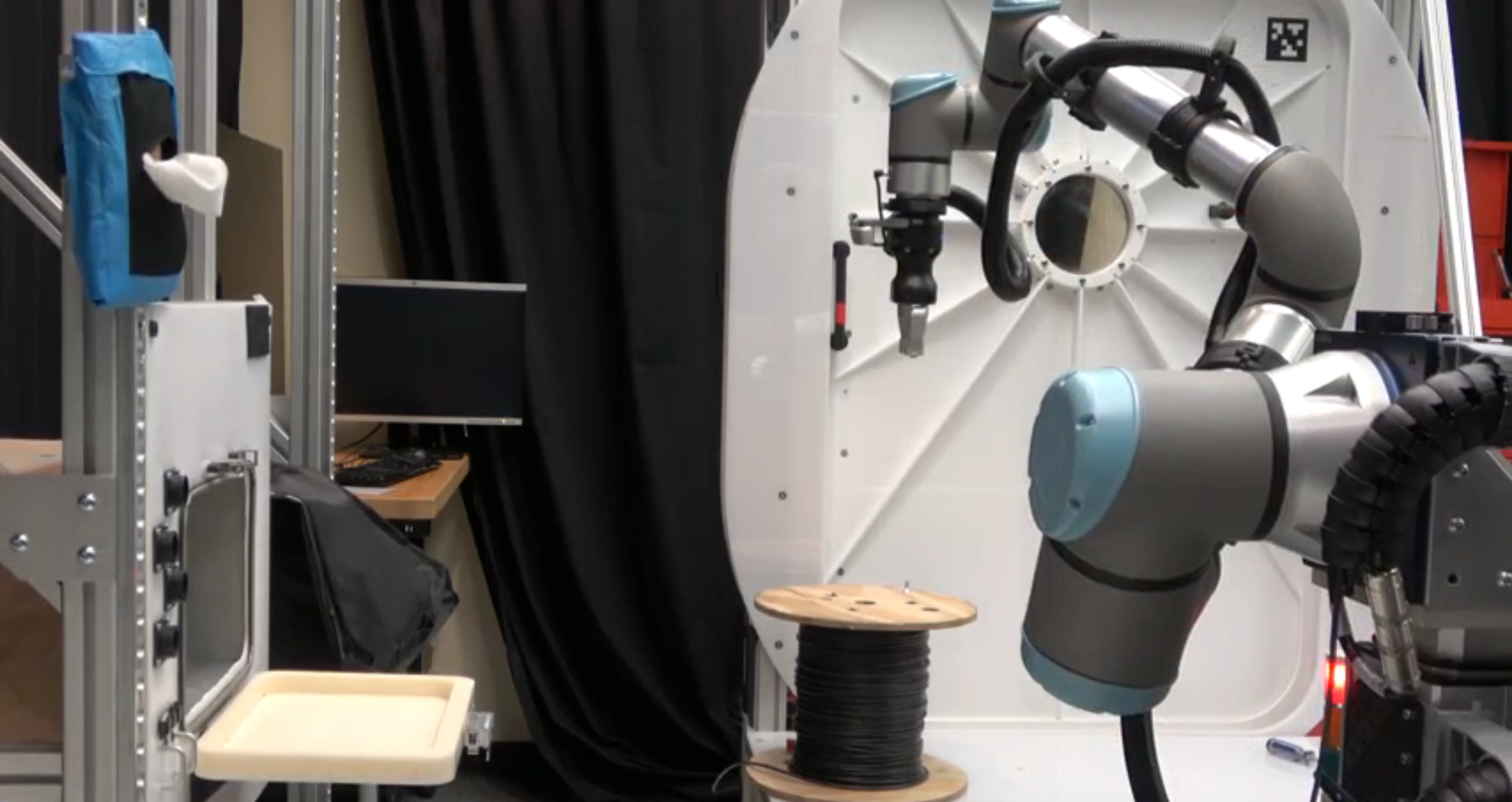}
        \caption{
        Robot requests help to proceed when the sample fell out of reach.
        % The robot asks for help to proceed when it does not detect the sample
        % % it expected to pick up 
        % because it fell out of reach.
        }
    \end{subfigure}
    
    \caption{Select trials of iMETRO performing a sample stowage task as if in a lunar habitat using \sar.}
    \label{fig:clr_trials}
\end{figure*}

\begin{figure*}
    \centering
    \begin{subfigure}{0.3\textwidth}
        \centering
        \includegraphics[height=3.5cm,keepaspectratio]{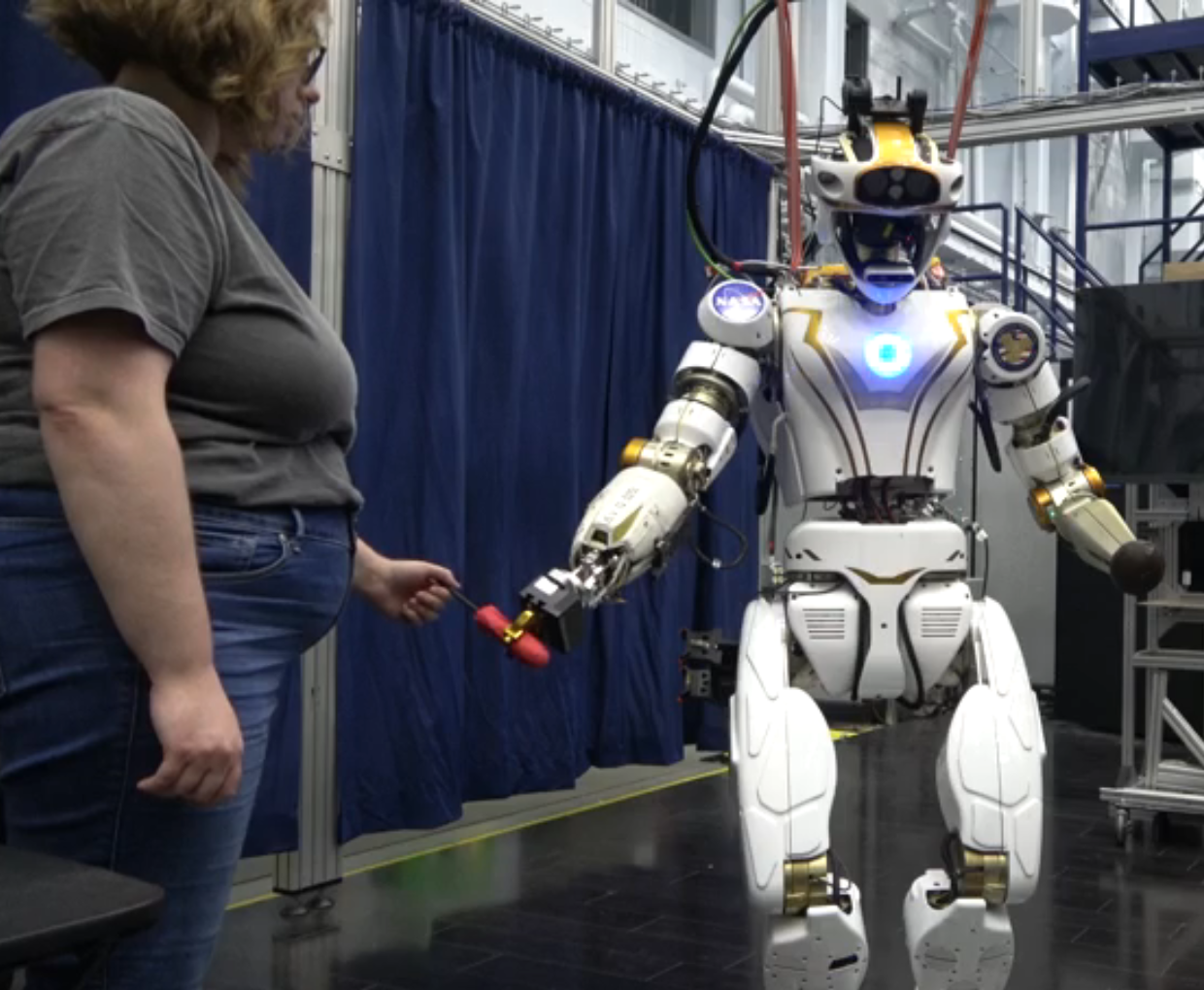}
        \caption{
        % Safe tool hand-off.
        Robot safely performs the tool hand-off task.
        }
    \end{subfigure}
    \hfill
    % \par\bigskip
    \begin{subfigure}{0.3\textwidth}
        \centering
        \includegraphics[height=3.5cm,keepaspectratio]{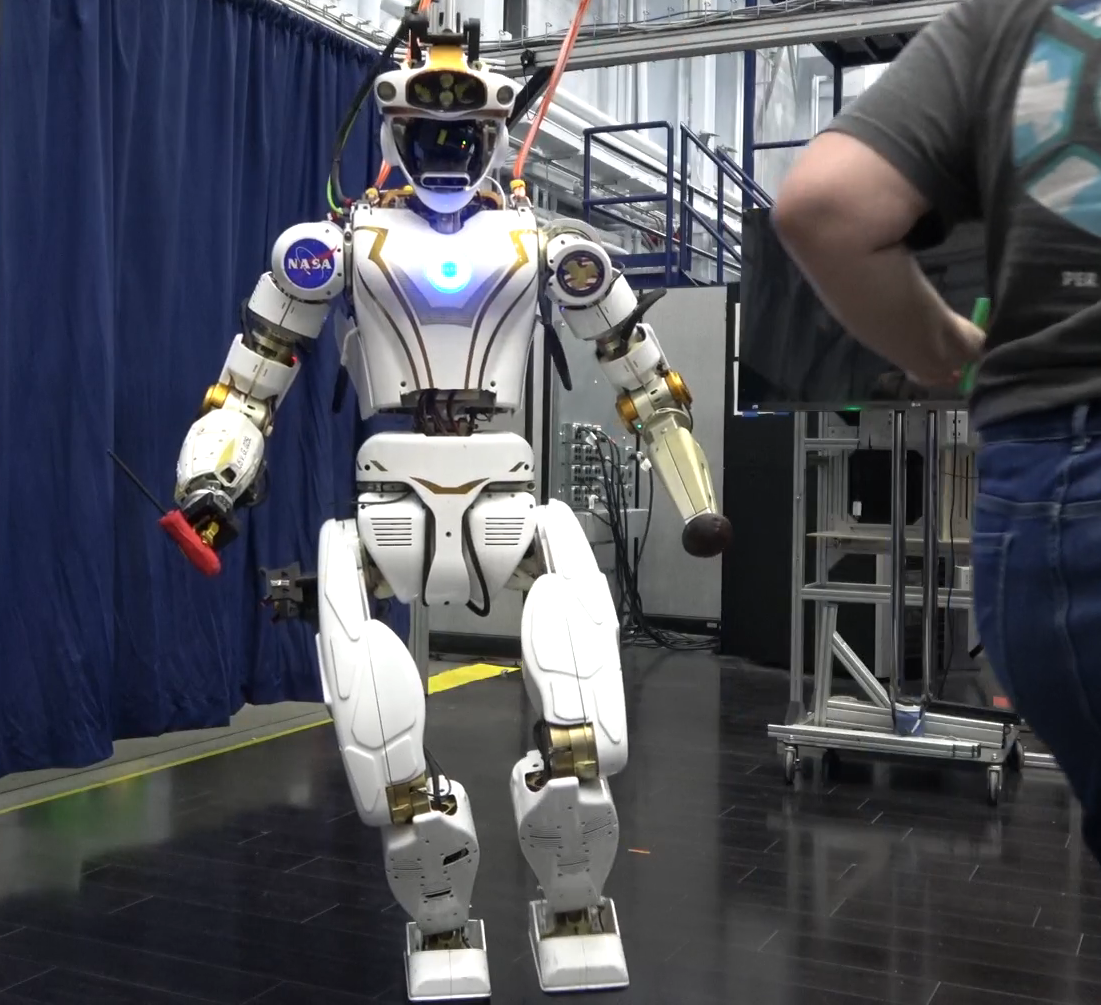}
        \caption{
        Robot moves slowly as human walks through workspace.
        % The robot moves more slowly because a human resident walks through the workspace.
        }
    \end{subfigure}
    \hfill
    % \par\bigskip
    \begin{subfigure}{0.3\textwidth}
        \centering
        \includegraphics[height=3.5cm,keepaspectratio]{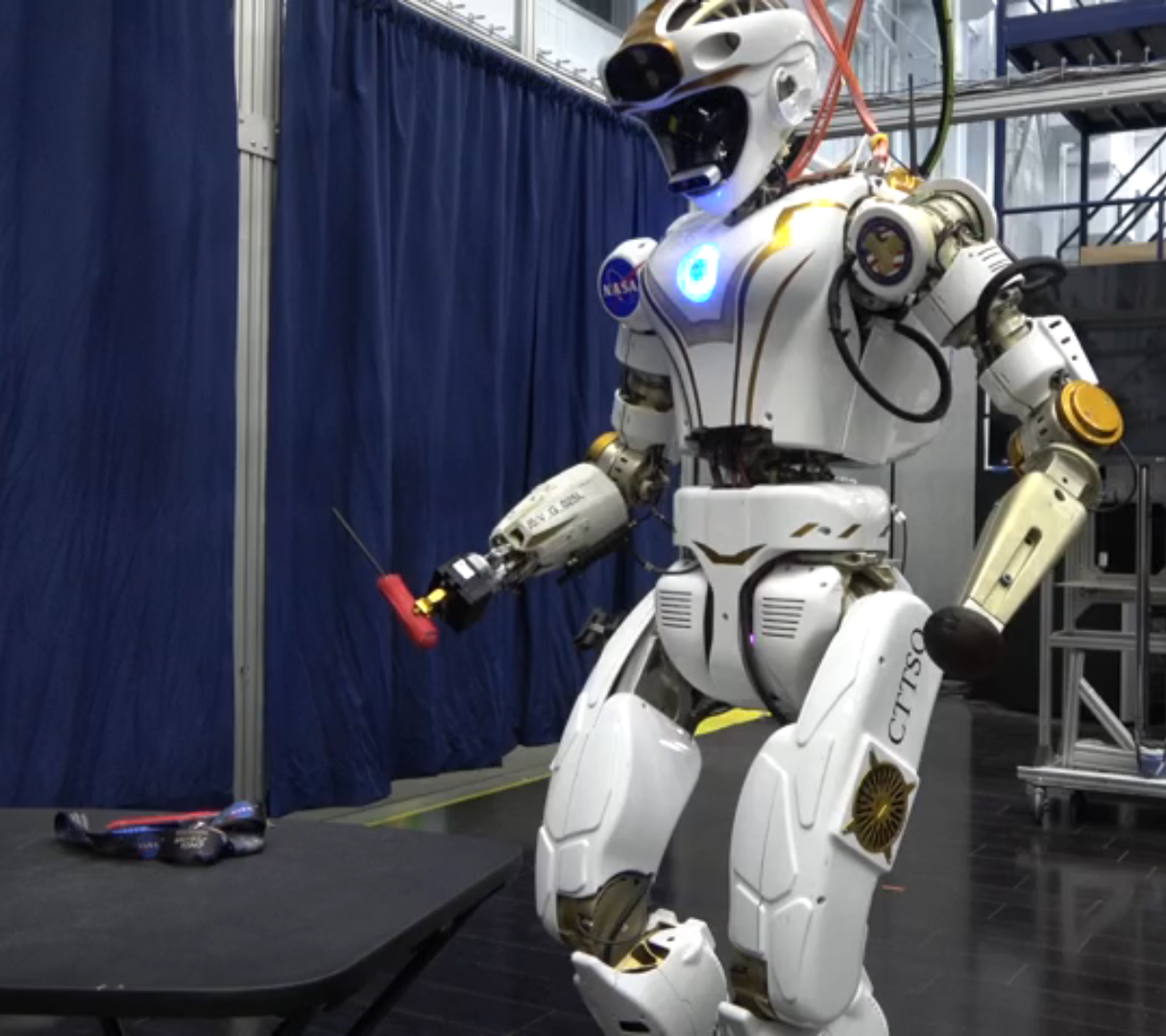}
        \caption{
        Robot requests supervision for possible collision with table.
        % The robot asks for supervision to proceed when it detects a possible collision with the objects on the table.
        }
    \end{subfigure}
    
    \caption{Select trials of Valkyrie performing a tool hand-off task as if in a household using \sar.}
    \label{fig:val_trials}
\end{figure*}

To investigate an example of how the \hrrt paradigm may function with different model representations, we aimed to test how robots can perform risk mitigating actions during task execution in two domains with different definitions of safety---lunar habitat and household.
We trained environment-specific logistic regression \rmaums on information learned from the \HRRTing exercise to identify appropriate risk mitigating actions during task execution.
% To test how the \HRRTing approach can help robots perform risk mitigation actions, we narrowed our scope to two domains with different definitions of safety---lunar habitat and household.
% In particular,
% we aim to model the \rmau of these two problem domains
% with different definitions of safety---lunar habitat and household---
% on two robots---
We considered two robots: NASA Johnson Space Center's iMETRO (Integrated Mobile Evaluation Testbed for Robotics Operations)~\cite{azimi2023imetro} %, imetropicknik}
and Valkyrie robot~\cite{Radford2015, IHMC2023}. % , Jorgensen2019, Jorgensen2022
% Valkyrie uses IHMC Open Robotics Software \cite{IHMC2023} for whole-body control.
Both Valkyrie and iMETRO use ROS2 \cite{ROS2} and MoveIt 2 \cite{MoveIt2_2024} %\cite{MoveIt2014, MoveIt2_2024}
for motion planning and execution.
Since Valkyrie (\fref{fig:val_trials}) is a legged humanoid robot, Valkyrie operates in terrestrial environments, such as households.
The iMETRO facility (\fref{fig:clr_trials}) was developed for testing capabilities required of assistive robots in lunar habitats, where confined spaces, dangerous environments, and high travel costs make operations significantly riskier.
We expect our \HRRTing approach to \sar to handle different environmental safety requirements as well as the different risk mitigating actions these robots are capable of performing.

\begin{table}
    \centering
    % INCLUDING TASK COMPLETION SUCCESS RATE
    % \begin{tabular}{|c|c|c|c|c|}
    %     \hline
    %     \multirow{2}{*}{\textbf{Environment}} & \multirow{2}{*}{\textbf{Robot}} & \multirow{2}{*}{\textbf{Total Trials}} & \textbf{Correct Risk Mitigating} & \textbf{Task Completion} \\
    %      & & & \textbf{Action Success Rate} & \textbf{Success Rate} \\
    %     \hline \hline
    %     Lunar Habitat & iMETRO & 7 & 1.00 & 0.43 \\
    %     \hline
    %     Household & Valkyrie & 5 & 0.60 & 0.60\\
    %     \hline \hline
    %     \textbf{Cumulative} & & 12 & \textbf{0.83} & 0.50 \\
    %     \hline
    % \end{tabular}

    \begin{tabular}{|c|c|c|c|}
        \hline
        \multirow{2}{*}{\textbf{Environment}} & \multirow{2}{*}{\textbf{Robot}} & \multirow{2}{*}{\textbf{Total Trials}} & \textbf{Risk Mitigation} \\
         & & & \textbf{Success Rate} \\
        \hline \hline
        Lunar Habitat & iMETRO & 7 & 1.00 \\
        \hline
        Household & Valkyrie & 5 & 0.60 \\
        \hline \hline
        \textbf{Cumulative} & - & 12 & \textbf{0.83} \\
        \hline
    \end{tabular}
    \caption{\Sar experiment results across 12 total trials.
    Errors in risk mitigation
    % identifying the appropriate risk mitigating action 
    are due to false negatives in hazard detection, namely our use of color blob detection \cite{colorblob} where lighting conditions impacted perception of color.
    When the robots correctly identified hazards,
    % the risk mitigating action, 
    they successfully mitigated risks to complete the task safely.
    % When the task could not be completed, it was because the robot failed to mitigate the risks, an object fell where it could not be recovered, or motion planning failures.
    }
    \label{table:results}
\end{table}

To detect hazards, we used color blob detection~\cite{colorblob} and YOLO object detection~\cite{YOLOpaper, YOLOcode}.
When any hazard was identified, %$\psi_t(s_t) = T$, 
the trained \rmaum
% for the environment was used to 
predicted the action-utilities of the risk mitigating actions and the action with the highest utility was executed.
% and the \rmp function $\pi_R$ (\eref{eq:action_prediction}) computed the best risk mitigating action $a_{R,t} = \pi_R(s_t) = \arg \max(\hat{U}_R(s_t,\Psi))$.
When no hazards were detected,
% When the total state risk $\Phi_R(s_t)$ was below risk threshold $R^*$,
the robot continued task execution. %executing task actions in $A_T$.
% We used an intentionally low risk threshold $R^* = 2.0$, which corresponded to the risk score of the risk condition with the lowest consequence severity.  This means that if any condition but the least consequential condition was present, the robot would have to mitigate the risks before continuing with the task.

The iMETRO robot performed 7 trials as if in a lunar habitat and Valkyrie performed 5 trials as if in a household.  In each trial, the robots were presented with different
% subsets of
hazards.  We recorded whether the robot identified and performed an appropriate risk mitigating action and whether the task was executed safely.
% , and whether the task was completely successfully.
The cumulative results
% our \rmaums 
can be seen in \tref{table:results}.
% See \aref{app:HRRT_experiments} for more detailed information about all \sar risk mitigation trials.
\fref{fig:clr_trials} and \fref{fig:val_trials} show select examples of the performed \sar tasks for the lunar habitat and household, respectively.
% It was also important for the robot to report the identified hazards and risk mitigating action it performed to the operator.
% Examples of the robots safety reports can be seen in \fref{fig:safety_reports}.

\subsection{\textit{Human-Robot Red Teaming} Results}
\label{sec:HRRT_experiments_results}

We evaluated how our \HRRTing methods achieved \sar in symbolic planning (\sref{subsec:HRRT_plan_exps}) and in robot execution (\sref{subsec:HRRT_robot_exps}) tasks.
This evaluation illustrates that the \hrrt can be applied to different types of problems that require \sar.
Taken together, our results demonstrate that the \HRRTing approach and iterations through the \HRRT levels improve the robot's ability to reason over and execute tasks in safety-critical domains.
Furthermore, the reflective cooperative nature of the \HRRTing exercise (especially through the English-like interactions on level HRRT4) has the potential to improve the combined human-robot team's understanding of the risks, critical failures, and complexities of the environment.

\section{DISCUSSION AND CONCLUSION}
\label{sec:disc_concl}

We demonstrate that the \hrrt paradigm can effectively inform \sar for safe planning in different problem domains and risk mitigation by robots with different embodiments acting in different environments.
% to appropriately assess and mitigate risks.
Using the \HRRTing approach, human-robot teams explore safety in the environment. % the space of hazards that could appear in an environment.
% challenged assumptions about the risks present in an environment and
We demonstrate that across 8 planning domains, the robot safely completes symbolic planning tasks with a success rate of 0.875.  Planning failures occurred due to the complexity of the explored domain, suggesting that more iterations through the \HRRT levels would eventually uncover the relevant information for safe planning.
We demonstrate how this domain exploration can inform risk assessment in physical robot execution experiments, specifically by training environment-specific \rmaums, which predict the action-utility of risk mitigating actions.
By selecting the action %.  The \rmp selects the action 
that would most effectively mitigate risks, the robot completes tasks safely with a success rate of 0.83.  Failures in identifying hazards occurred due to false negatives from our perception modules, highlighting the need for further work in effective hazard identification.

Future work includes investigating the composition of human-robot teams, specifically by recruiting independent expert humans to provide insights into the problem.
Performing similar robot task execution experiments after successive model iterations, in more problem domains, and in more evaluation tasks per domain would provide additional insight into how the \HRRT iterative methods translate to robot hardware.
% online learning of the hazard likelihood functions, risk mitigating action space, and safe task performance.
% the hazard likelihood functions rather than relying on human operators to provide these function.
% We would also want the robot to perform lifelong learning about safe task performance, which would help correct mistakes more quickly.
% Expanding the NLP capabilities would enable more intuitive interaction on human-robot teams.
% Counter-factual reasoning---either in simulation or during task execution---would also help the robot identify new hazard conditions, consequences, and risk mitigating actions.
% Finally, w
Addressing perceptual challenges in hazard identification, specifically differentiating between safe and unsafe operating conditions, will require additional research.

% We propose the \hrrt paradigm for \sar and demonstrate the feasibility of this approach to mitigating risks in safety-critical domains.
% Through the \hrrt, human and robot agents work together to challenge assumptions and explore the state of possible hazardous situations the robot may face.
% We use this exploration to generate \HRRTed datasets to train \rmaums, which evaluate the state-action utility values of the risk mitigating actions.
We suggest that the \HRRTing paradigm for \sar deserves further study and broader application based on the promise demonstrated in this paper.
Even with the simple English-like interactions carried out by our \HRRT implementation, our symbolic planning and robot execution experiments demonstrate that useful information is gained from the collaborative process of challenging and reflecting on the team's modeled knowledge.
Our work demonstrates that robots with different embodiments can effectively and safely plan and operate in different environments under different definitions of safety, helping robots to earn trust as collaborators in safety-critical tasks.

% \todo{draft this section}

% Future work:
% \begin{itemize}
%     \item consider $MAS$ in terms of cooperative human/robot teams, but could be extended to include other agents such as adversaries actively trying to force the robot into performing unsafe actions
%     \item learn likelihood function
%     \item expand NLP capabilities: Note that while we use language as an intuitive interface for communicating task goals to robots, the focus of this project is on the \textit{safe reasoning over} and \textit{safe execution} of those commands.  We take inspiration from DOROTHIE \cite{DOROTHIE2022} and implement an ``English-like'' interface in the form of a dialogue decision tree, rather than a complete natural language interface.  The focus of our work is to enable human and robot to engage in dialogue together about safety.  As some natural language capabilities are outside the scope of this project, we leave it to future work to fully explore the ways in which a similar safety-aware natural language interface may work and the abundance of commands that could be processed by this interface.
%     \item online refinement/learning after model deployment; \hrrt can perform counter-factual analysis of \sar task execution to identify alternate risk mitigating actions that could have been taken instead
% \end{itemize}

\section*{Acknowledgments}

This work was supported in part by %NASA NSTGRO
NASA Space Technology Graduate Research Opportunity (NSTGRO)
grant 80NSSC20K1200.
We would like to thank the members of the NASA Johnson Space Center Dexterous Robotics Team.
Special thanks to Mina Kian for her appearance in experiment photos and videos.
% Special thanks to Steven Jens Jorgensen, Misha Savchenko, Mark Paterson, Ian Chase, Lewis Hill, and Mina Kian for their mentorship, input, and robot ops support.

% %%%%%%%%%%%%%%%%%%%%%%%%%%%%%%%%%%%%%%%%%%%%%%%%%%%%%%%%%%%%%%%%%%%%%%%%%%%%%%%%

% References are important to the reader; therefore, each citation must be complete and correct. If at all possible, references should be commonly available publications.


\begin{thebibliography}{99}

% TODO UPDATE BIBLIGRAPHY, CHECK FORMATTING
% TODO ALL REFERENCES FORMATTED CONSISTENTLY
% TODO ADD REFERENCES TO ANNOTATED BIBLIOGRAPHY

\bibitem{SpaceOdyssey} \textit{2001: A Space Odyssey}, Directed by Stanley Kubrick, Stanley Kubrick Productions, 1968.

% A----------

\bibitem{Abbass2011} H. Abbass, A. Bender, S. Gaidow, and P. Whitbread, ``Computational Red Teaming: Past, Present, and Future,'' \textit{IEEE Computational Intelligence Magazine}, 2011.

\bibitem{Adkins2013} G. Adkins, ``Red Teaming the Red Team: Utilizing Cyber Espionage to Combat Terrorism,'' \textit{Journal of Strategic Security}, 2013.

% \bibitem{SayCan2022} M. Ahn, A. Brohan, N. Brown, Y. Chebotar, O. Cortes, B. David, C. Finn, C. Fu, K. Gopalakrishnan, K. Hausman, A. Herzog, D. Ho, J. Hsu, J. Ibarz, B. Ichter, A. Irpan, E. Jang, R. J. Ruano, K. Jeffrey, S. Jesmonth, N. Joshi, R. Julian, D. Kalashnikov, Y. Kuang, K.-H. Lee, S. Levine, Y. Lu, L. Luu, C. Parada, P. Pastor, J. Quiambao, K. Rao, J. Rettinghouse, D. Reyes, P. Sermanet, N. Sievers, C. Tan, A. Toshev, V. Vanhoucke, F. Xia, T. Xiao, P. Xu, S. Xu, M. Yan, and A. Zeng, ``Do As I Can, Not As I Say: Grounding Language in Robotic Affordances,'' \textit{arXiv preprint arXiv:2204.01691}, 2022.

% \bibitem{Allott2005} N. Allott, ``Paul Grice, Reasoning and Pragmatics,'' \textit{UCL Working Papers in Linguistics}, 2005.

\bibitem{Pyperplan} Y. Alkhazraji, M. Frorath, M. Gr\"{u}tzner, M. Helmert, T. Liebetraut, R. Mattm\"{u}ller, M. Ortlieb, J. Seipp, T. Springenberg, P. Stahl, and J. W\"{u}lfing, ``Pyperplan,'' 2020. [Online]. Available: \url{https://doi.org/10.5281/zenodo.3700819}

% \bibitem{Argall2009} B. D. Argall, S. Chernova, M. Veloso, and B. Browning, ``A Survey of Robot Learning from Demonstration,'' \textit{Robotics and Autonomous Systems}, 2009.

% \bibitem{Atienza2002} R. Atienza and A. Zelinsky, ``Active Gaze Tracking for Human-Robot Interaction,'' \textit{IEEE International Conference on Multimodal Interfaces}, 2002.

\bibitem{azimi2023imetro} S. Azimi, ``iMETRO (Integrated Mobile Evaluation Testbed for Robotics Operations) Facility,'' NASA Technical Reports Server (NTRS), 2023. [Online]. Available: \url{https://ntrs.nasa.gov/citations/20230015485}

% B----------

% \bibitem{Beeson2010} P. Beeson, J. Modayil, and B. Kuipers, ``Factoring the Mapping Problem: Mobile Robot Map-Building in the Hybrid Spatial Semantic Hierarchy,'' \textit{The International Journal of Robotics Research}, 2010.

% \bibitem{SafetyHouseCleaning} J. Bennett, ``9 Cleaning Mistakes That Are Making Your Home Dirtier,'' Better Homes and Gardens, 2024. [Online]. Available: \url{https://www.bhg.com/homekeeping/house-cleaning/tips/cleaning-mistakes/}

\bibitem{IHMC2023} S. Bertrand, D. Calvert, S. McCrory, R. Griffin, B. Mishra, J. Foster, D. Anderson, L. Penco, and N. Kitchel, ``IHMC Open Robotics Software,'' 2023. [Online]. Available: \url{https://github.com/ihmcrobotics/ihmc-open-robotics-software}

\bibitem{Billard2019} A. Billard and D. Kragic, ``Trends and Challenges in Robot Manipulation,'' \textit{Science}, 2019.

\bibitem{Bogue2017} R. Bogue, ``Robots that Interact with Humans: A Review of Safety Technologies and Standards,'' \textit{Industrial Robot: An International Journal}, 2017.

\bibitem{Bozhinoski2019} D. Bozhinoski, D. Di Ruscio, I. Malavolta, P. Pelliccione, and I. Crnkovic, ``Safety for Mobile Robotic Systems: A Systematic Mapping Study from a Software Engineering Perspective,'' \textit{Journal of Systems and Software}, 2019.

% \bibitem{Bruce1993} V. Bruce, ``What the Human Face Tells the Human Mind: Some Challenges for the Robot-Human Interface,'' \textit{Advanced Robotics}, 1993.

% C----------

% \bibitem{SafetyAI_IBM} R. D. Caballar, ``10 AI Dangers and Risks and How to Manage Them,'' IBM, 2024. [Online]. Available: \url{https://www.ibm.com/think/insights/10-ai-dangers-and-risks-and-how-to-manage-them}

\bibitem{SafetyAI} Center for AI Safety, ``An Overview of Catastrophic AI Risks,'' Center for AI Safety, 2024. [Online]. Available: \url{https://www.safe.ai/ai-risk}

\bibitem{SafetyNuclear} Centers for Disease Control and Prevention (CDC), ``Nuclear Weapon Infographic,'' Centers for Disease Control and Prevention (CDC) radiation Emergencies, 2024. [Online]. Available: \url{https://www.cdc.gov/radiation-emergencies/infographic/nuclear-weapon.html}

% \bibitem{DQN2019} Y. Chen and E. Kulla, ``A Deep Q-Network with Experience Optimization (DQN-EO) for Atari's Space Invaders,'' \textit{Workshops of the International Conference on Advanced Information Networking and Applications (WAINA)}, 2019.

\bibitem{Chen2022} Y. Chen, C. Yang, Y. Gu, and B. Hu, ``Influence of Mobile Robots on Human Safety Perception and System Productivity in Wholesale and Retail Trade Environments: A Pilot Study,'' \textit{IEEE Transactions on Human-Machine Systems}, 2022.

% \bibitem{Chernova2010} S. Chernova, J. Orkin, and C. Breazeal, ``Crowdsourcing HRI through Online Multiplayer Games,'' \textit{AAAI Fall Symposium Series}, 2010.

% \bibitem{RL2020Survey} J. Clifton and E. Laber, ``Q-Learning: Theory and Applications,'' \textit{Annual Review of Statistics and Its Application}, 2020.

% \bibitem{MoveIt2014} D. Coleman, I. Sucan, S. Chitta, and N. Correll, ``Reducing the Barrier to Entry of Complex Robotic Software: A MoveIt! Case Study,'' \textit{arXiv preprint arXiv:1404.3785}, 2014.

% \bibitem{Cowan2017} B. R. Cowan, N. Pantidi, D. Coyle, K. Morrissey, P. Clarke, S. Al-Shehri, D. Earley, and N. Bandeira, ````What Can I Help You With?'': Infrequent Users' Experiences of Intelligent Personal Assistants,'' \textit{International Conference on Human-Computer Interaction with Mobile Devices and Services}, 2017.

% D----------

\bibitem{Dhillon1993} B. S. Dhillon and O. C. Anude, ``Robot Safety and Reliability: A Review,'' \textit{Microelectronics Reliability}, 1993.

\bibitem{Dhillon2002} B. S. Dhillon, A. R. M. Fashandi, and K. L. Liu, ``Robot Systems Reliability and Safety: A Review,'' \textit{Journal of Quality in Maintenance Engineering}, 2002.

% \bibitem{Howard2016inferring} F. Duvallet, M. R. Walter, T. Howard, S. Hemachandra, J. Oh, S. Teller, N. Roy, and A. Stentz, ``Inferring Maps and Behaviors from Natural Language Instructions,'' \textit{Experimental Robotics, Spring International Publishing}, 2016.

% \bibitem{Dzifcak2009} J. Dzifcak, M. Scheutz, C. Baral, and P. Schermerhorn, ``What to Do and How to Do It: Translating Natural Language Directives Into Temporal and Dynamic Logic Representation for Goal Management and Action Execution,'' \textit{IEEE International Conference on Robotics and Automation}, 2009.

% E----------

% \bibitem{CIMONBlog} T. Eisenberg, ``CIMON, the AI-Powered Robot, Launches a New Era in Space Travel,'' IBM, 2019. [Online]. Available: \url{https://www.ibm.com/blog/cimon-ai-robot-launches-new-era-space-travel/}

% \bibitem{Roese2008} K. Epstude and N. J. Roese, ``The Functional Theory of Counterfactual Thinking,'' \textit{Personality and Social Psychology Review}, 2008.

\bibitem{Roese2011} K. Epstude and N. J. Roese, ``When Goal Pursuit Fails: The Functions of Counterfactual Thought in Intention Formation,'' \textit{Social Psychology}, 2011.

% F----------

% \bibitem{SafetyTravel} A. Fraiel, ``10 Common Travel Problems and How to Deal with Them,'' Worldpackers, 2025. [Online]. Available: \url{https://www.worldpackers.com/articles/common-travel-problems-and-how-to-deal-with-them}

% G----------

\bibitem{Ganguli2022} D. Ganguli, L. Lovitt, J. Kernion, A. Askell, Y. Bai, S. Kadavath, B. Mann, E. Perez, N. Schiefer, K. Ndousse, A. Jones, S. Bowman, A. Chen, T. Conerly, N. DasSarma, D. Drain, N. Elhage, S. El-Showk, S. Fort, Z. Hatfield-Dodds, T. Henighan, D. Hernandez, T. Hume, J. Jacobson, S. Johnston, S. Kravec, C. Olsson, S. Ringer, E. Tran-Johnson, D. Amodei, T. Brown, N. Joseph, S. McCandlish, C. Olah, J. Kaplan, and J. Clark, ``Red Teaming Language Models to Reduce Harms: Methods, Scaling Behaviors, and Lessons Learned,'' \textit{arXiv preprint arXiv:2209.07858}, 2022.

\bibitem{PDDL1998} M. Ghallab, A. Howe, C. Knoblock, D. McDermott, A. Ram, M. Veloso, D. Weld, and D. Wilkins, ``PDDL---The Planning Domain Definition Language,'' \textit{Technical Report}, 1998.

% \bibitem{FMEA_software} P. L. Goddard, ``Software FMEA Techniques,'' \textit{Annual Reliability and Maintainability Symposium, International Symposium on Product Quality and Integration}, 2000.

% \bibitem{Grice1975} H. P. Grice, ``Logice and Conversation,'' \textit{Speach Acts}, 1975.

% \bibitem{Grice1991} P. Grice, \textit{Studies in the Way of Words}, Harvard University Press, 1991.

\bibitem{Guevara2024} P. Guevara, ``A Guide to Understanding 5x5 Risk Assessment Matrix,'' SafetyCulture, 2024. [Online]. Available: \url{https://safetyculture.com/topics/risk-assessment/5x5-risk-matrix/}

\bibitem{Guiochet2017} J. Guiochet, M. Machin, and H. Waeselynck, ``Safety-Critical Advanced Robots: A Survey,'' \textit{Robotics and Autonomous Systems}, 2017.

% H----------

% \bibitem{Hambuchen2004} K. Hambuchen, ``Multi-Modal Attention and Event Binding in Humanoid Robot Using a Sensing Ego-Sphere,'' Ph.D. Dissertation, Vanderbilt University, 2004.

% \bibitem{Howard2015} S. Hemachandra, F. Duvallet, T. M. Howard, N. Roy, A. Stentz, M. R. Walter, ``Learning Models for Following Natural Language Directions in Unknown Environments,'' \textit{IEEE International Conference on Robotics and Automation (ICRA)}, 2015.

\bibitem{Hentout2019} A. Hentout, M. Aouache, A. Maoudj, and I. Akli, ``Human-Robot Interaction in Industrial Collaborative Robotics: A Literature Review of the Decade 2008-2017,'' \textit{Advanced Robotics}, 2019.

% \bibitem{Hermann2014} K. M. Hermann, D. Das, J. Weston, and K. Ganchev, ``Semantic Frame Identification with Distributed Word Representations,'' \textit{Association for Computational Linguistics}, 2014.

\bibitem{Hopko2024} S. K. Hopko and R. K. Mehta, ``Trust in Shared-Space Collaborative Robots: Shedding Light on the Human Brain,'' \textit{Human Factors}, 2024.

% \bibitem{Howard2014} T. M. Howard, S. Tellex, and N. Roy, ``A Natural Language Planner Interface for Mobile Manipulators,'' \textit{IEEE International Conference on Robotics and Automation (ICRA)}, 2014.

% I----------

% \bibitem{CIMONWatson} IBM, ``CIMON: IBM Watson in Space,'' IBM, 2018. [Online]. Available: \url{https://www.ibm.com/thought-leadership/smart/ai-in-space-xp/journal.html}

% \bibitem{CIMON} IBM, ``CIMON Brings AI to the International Space Station,'' IBM, 2019. [Online]. Available: \url{https://www.ibm.com/thought-leadership/innovation-explanations/cimon-ai-in-space}

\bibitem{IronGiant} \textit{The Iron Giant}, Directed by Brad Bird, Warner Bros., 1999.

% J----------

% \bibitem{RL2019} B. Jang, M. Kim, G. Harerimana, and J. W. Kim, ``Q-Learning Algorithms: A Comprehensive Classification and Applications,'' \textit{IEEE Access}, 2019.

\bibitem{YOLOcode} G. Jocher \textit{et al.}, ``ultralytics/yolov5: v7.0 - YOLOv5 SOTA Realtime Instance Segmentation,'' \textit{Zenodo}, November 22, 2022. doi: 10.5281/zenodo.7347926.

% \bibitem{Jorgensen2019} S. J. Jorgensen, M. W. Lanighan, S. S. Bertrand, A. Watson, J. S. Altemus, R. S. Askew, L. Bridgwater, B. Domingue, C. Kendrick, J. Lee, M. Paterson, J. Sanchez, P. Beeson, S. Gee, S. Hart, A. H. Quispe, R. Griffin, I. Lee, S. McCrory, L. Sentis, J. Pratt, and J. S. Mehling, ``Deploying the NASA Valkyrie Humanoid for IED Response: An Initial Approach and Evaluation Summary,'' \textit{IEEE-RAS International Conference on Humanoid Robots (Humanoids)}, 2019.

% \bibitem{Jorgensen2022} S. J. Jorgensen, M. Wonsick, M. Paterson, A. Watson, I. Chase, and J. S. Mehling, ``Cockpit Interface for Locomotion and Manipulation Control of the NASA Valkyrie Humanoid in Virtual Reality (VR),'' NASA Technical Reports Server, 2022. [Online]. Available: \url{https://ntrs.nasa.gov/citations/20220007587}

% \bibitem{NASAIncidents} JSC Safety and Mission Assurance (SMA) Flight Safety Office, ``Significant Incidents and Close Calls in Human Spaceflight,'' NASA Safety and Mission Assurance (SMA), 2024. [Online]. Available: \url{https://sma.nasa.gov/SignificantIncidents/}

% K----------

% \bibitem{Keenan1976} E. O. Keenan, ``The Universality of Conversational Postulates,'' \textit{Language in Society}, 1976.

% \bibitem{king2001logistic} G. King and L. Zeng, ``Logistic Regression in Rare Events Data,'' \textit{Political Analysis}, 2001.

% \bibitem{Klein2003} D. Klein and C. D. Manning, ``Accurate Unlexicalized Parsing,'' \textit{Association for Computational Linguistics}, 2003.

% \bibitem{Koidan2019} K. Koidan, ``7 Effective Ways to Deal With a Small Dataset,'' \textit{Hackernoon}, 2019. [Online]. Available: \url{https://hackernoon.com/7-effective-ways-to-deal-with-a-small-dataset-2gyl407s}

% \bibitem{Kollar2010} T. Kollar, S. Tellex, D. Roy, and N. Roy, ``Toward Understanding Natural Language Directions,'' \textit{ACM/IEEE International Conference on Human-Robot Interaction (HRI)}, 2010.

% \bibitem{Kollar2010grounding} T. Kollar, S. Tellex, D. Roy, N. Roy, ``Grounding Verbs of Motion in Natural Language Commands to Robots,'' \textit{International Symposium on Experimental Robotics (ISER)}, 2014.

% \bibitem{Konidaris2012} G. Konidaris, S. Kuindersma, R. Grupen, and A. Barto, ``Robot Learning from Demonstration by Constructing Skill Trees,'' \textit{International Journal of Robotics Research}, 2012.

\bibitem{Kraemer2004} S. Kraemer, P. Carayon, and R. Duggan, ``Red Team Performance for Improved Computer Security,'' \textit{Proceedings of the Human Factors and Ergonomics Society Annual Meeting}, 2004.

% \bibitem{Kuipers2000} B. Kuipers, ``The Spatial Semantic Hierarchy,'' \textit{Artificial Intelligence}, 2000.

\bibitem{Kuipers2018} B. Kuipers, ``How Can We Trust a Robot?,'' \textit{Communications of the ACM}, 61(3):86-95, 2018.

\bibitem{Kuipers2020} B. Kuipers, ``Perspectives on Ethics of AI,'' in \textit{The Oxford Handbook of Ethics of AI}, pages 421-441, Oxford University Press, 2020.

\bibitem{Kuipers2022} B. Kuipers, ``Trust and Cooperation,'' \textit{Frontiers in Robotics and AI}, 9:676767, 2022.

\bibitem{Kuipers2023} B. Kuipers, ``AI and Society: Ethics, Trust, and Cooperation,'' \textit{Communications of the ACM}, 66(8):35-38, 2023.

% \bibitem{RL2020} A. Kumar, A. Zhou, G. Tucker, and S. Levine, ``Conservative Q-Learning for Offline Reinforcement Learning,'' \textit{Advances in Neural Information Processing Systems (NEURIPS)}, 2020.

% L----------

% \bibitem{lane2001toward} T. Lane and L. P. Kaelbling, ``Toward Hierarchical Decomposition for Planning in Uncertain Environments,'' \textit{IJCAI Workshop on Planning Under Uncertainty and Incomplete Information}, 2001.

\bibitem{Lasota2017} P. A. Lasota, T. Fong, and J. A. Shah, ``A Survey of Methods for Safe Human-Robot Interaction,'' \textit{Foundations and Trends in Robotics}, 2017.

\bibitem{Lee2004} J. D. Lee and K. A. See, ``Trust in Automation: Designing for Appropriate Reliance,'' \textit{Human Factors}, 2004.

\bibitem{Lim2000} H. O. Lim and K. Tanie, ``Human Safety Mechanisms of Human-Friendly Robots: Passive Viscoelastic Trunk and Passively Movable Base,'' \textit{The International Journal of Robotics Research}, 2000.

\bibitem{Littman2013} M. L. Littman, T. L. Dean, and L. P. Kaelbling, ``On the Complexity of Solving Markov Decision Problems,'' \textit{arXiv preprint arXiv:1302.4971}, 2013.

\bibitem{Liu2020} Z. Liu, X. Wang, Y. Cai, W. Xu, Q. Liu, Z. Zhou, and D. T. Pham, ``Dynamic Risk Assessment and Active Response Strategy for Industrial Human-Robot Collaboration,'' \textit{Computers and Industrial Engineering}, 2020.

\bibitem{Longbine2008} D. F. Longbine, ``Red Teaming: Past and Present,'' \textit{School of Advanced Military Studies, Army Command and General Staff College}, 2008.

% \bibitem{Luger2016} E. Luger and A. Sellen, ````Like Having a Really Bad PA'': The Gulf between User Expectation and Experience of Conversational Agents,'' \textit{CHI Conference on Human Factors in Computing Systems}, 2016.

% M----------

\bibitem{DOROTHIE2022} Z. Ma, B. VanDerPloeg, C. P. Bara, H. Yidong, E. I. Kim, F. Gervits, M. Marg, and J. Chai, ``DOROTHIE: Spoken Dialogue for Handling Unexpected Situations in Interactive Autonomous Driving Agents,'' \textit{arXiv preprint arXiv:2210.12511}, 2022.

\bibitem{ROS2} S. Macenski, T. Foote, B. Gerkey, C. Lalancette, and W. Woodall, ``Robot Operating System 2: Design, Architecture, and Uses in the Wild,'' \textit{Science Robotics}, 2022.

% \bibitem{MacMahon2006} M. MacMahon, B. Stankiewicz, and B. Kuipers, ``Walk the Talk: Connecting Language, Knowledge, and Action in Route Instructions,'' \textit{AAAI}, 2006.

\bibitem{Mansfield2018} S. Mansfield-Devine, ``The Best Form of Defense---The Benefits of Red Teaming,'' \textit{Computer Fraud \& Security}, 2018.

% \bibitem{Matuszek2010directions} C. Matuszek, D. Fox, and K. Koscher, ``Following Directions Using Statistical Machine Translation,'' \textit{ACM/IEEE International Conference on Human-Robot Interaction (HRI)}, 2010.

% \bibitem{Matuszek2013parse} C. Matuszek, E. Herbst, L. Zettlemoyer, and D. Fox, ``Learning to Parse Natural Language Commands to a Robot Control System,'' \textit{Experimental Robotics}, 2013.

% \bibitem{Matuszek2014deictic} C. Matuszek, L. Bo, L. Zettlemoyer, and D. Fox, ``Learning from Unscripted Deictic Gesture and Language for Human-Robot Interactions,'' \textit{AAAI Conference on Artificial Intelligence}, 2014.

% \bibitem{Mnih2013DQN} V. Mnih, K. Kavukcuoglu, D. Silver, A. Graves, A. Ioannis, D. Wierstra, and M. Riedmiller, ``Playing Atari with Deep Reinforcement Learning,'' \textit{arXiv preprint arXiv:1312.5602}, 2013.

% \bibitem{Mnih2015DQN} V. Mnih, K. Kavukcuoglu, D. Silver, A. A. Rusu, J. Veness, M. G. Bellemare, A. Graves, M. Riedmiller, A. K. Fidjeland, G. Ostrovski, S. Petersen, C. Beattie, A. Sadik, I. Antonoglou, H. King, D. Kumaran, D. Wierstra, S. Legg, and D. Hassabis, ``Human-Level Control through Deep Reinforcement Learning,'' \textit{Nature}, 2015.

% \bibitem{Mu2019} J. Mu and A. Sarkar, ``Do We Need Natural Language? Exploring Restricted Language Interfaces for Complex Domains,'' \textit{CHI Conference on Human Factors in Computing Systems}, 2019.

% N----------

% \bibitem{NASAFMECA} NASA Glenn Research Center, ``Hazards Analysis and Failure Modes and Effects Criticality Analysis (FMECA) of Four Concept Vehicle Propulsion Systems,'' NASA/CR---2019-220217, 2019. [Online]. Available: \url{https://rotorcraft.arc.nasa.gov/Publications/files/CR-2019-220217.pdf}

\bibitem{NASAFMEA} NASA Goddard Space Flight Center, ``Guideline for Failure Modes and Effects Analysis and Risk Assessment,'' Goddard Technical Handbook 8004, NASA Goddard Space Flight Center, 2024. [Online]. Available: \url{https://standards.nasa.gov/sites/default/files/standards/GSFC/Baseline/0/GSFC-HDBK-8004_Approved_1.pdf}

\bibitem{NASANPR} NASA Office of Safety and Mission Assurance, ``NASA General Safety Program Requirements,'' NPR 8715.3, NASA, 2021.

\bibitem{NASAJPR} NASA Safety and Test Operations Division, ``JSC Safety and Health Requirements,'' JPR 1700.1, NASA Johnson Space Center, 2018. [Online]. Available: \url{https://www.nasa.gov/johnson/jsc-safety-health-requirements/}

\bibitem{SafetySpace} National Aeronautics and Space Administration, ``5 Hazards of Human Spaceflight,'' NASA, 2024. [Online]. Available: \url{https://www.nasa.gov/hrp/hazards/}

% \bibitem{Moon2Mars} National Aeronautics and Space Administration, ``Moon to Mars Strategy and Objectives,'' NASA, 2022. [Online]. Available: \url{https://www.nasa.gov/moontomarsarchitecture-strategyandobjectives/}

\bibitem{NASARiskManagement} National Aeronautics and Space Administration, ``NASA Risk Management Handbook,'' NASA Technical Reports Server, 2011. [Online]. Available: \url{https://ntrs.nasa.gov/api/citations/20120000033/downloads/20120000033.pdf}

\bibitem{NASASafety} National Aeronautics and Space Administration, ``NASA Safety Culture Handbook,'' NASA Technical Standards System, 2015. [Online]. Available: \url{https://standards.nasa.gov/sites/default/files/standards/NASA/Baseline/1/nasa-hdbk-870924_with_change_1.pdf}

% \bibitem{NASARiskManagementWebsite} National Aeronatutics and Space Administration, ``Risk Management,'' sma.nasa.gov, 2023. [Online]. Available: \url{https://sma.nasa.gov/sma-disciplines/risk-management}

% \bibitem{NASASafetyWebsite} National Aeronautics and Space Administration, ``Safety Culture,'' sma.nasa.gov, 2023. [Online]. Available: \url{https://sma.nasa.gov/sma-disciplines/safety-culture}

% \bibitem{Neale1992} S. Neale, ``Paul Grice and the Philosophy of Language,'' \textit{Linguistics and Philosophy}, 1992.

% \bibitem{Niekum2012} S. Niekum, S. Osentoski, G. Konidaris, and A. G. Barto, ``Learning and Generalization of Complex Tasks from Unstructured Demonstrations,'' \textit{IEEE International Conference on Intelligent Robots and Systems}, 2012.

% \bibitem{Niekum2013} S. Niekum, S. Chitta, A. G. Barto, B. Narthi, and S. Osentoski, ``Incremental Semantically Grounded Learning from Demonstration,'' \textit{Robotics: Science and Systems (RSS)}, 2013.

% O----------

\bibitem{OSHA_RCA} Occupational Safety and Health Administration, ``The Importance of Root Cause Analysis During Incident Investigation,'' OSHA Fact Sheet, 2016. [Online]. Available: \url{https://www.osha.gov/sites/default/files/publications/OSHA3895.pdf}

% \bibitem{DQN2015} J. Oh, X. Guo, H. Lee, R. Lewis, and S. Singh, ``Action-Conditional Video Prediction Using Deep Networks in Atari Games,'' \textit{Advances in Neural Information Processing Systems (NEURIPS)}, 2015.

\bibitem{ChatGPT} OpenAI, ChatGPT, 2025. Available: \url{https://chatgpt.com/}

% P----------

% \bibitem{Palinko2015} O. Palinko, F. Rea, G. Sandini, and A. Sciutti, ``Eye Gaze Tracking for a Humanoid Robot,'' \textit{IEEE-RAS International Conference on Humanoid Robots (Humanoids)}, 2015.

% \bibitem{Howard2016spatial} R. Paul, J. Arkin, N. Roy, and T. M. Howard, ``Efficient Grounding of Abstract Spatial Concepts for Natural Language Interaction with Robot Manipulators,'' \textit{Robotics: Science and Systems (RSS)}, 2016.

% \bibitem{Peace2017} C. Peace, ``The Risk Matrix: Uncertain Results?,'' \textit{Policy and Practice in Health and Safety}, 2017.

% \bibitem{Pearl1995causaldiagrams} J. Pearl, ``Causal Diagrams for Empirical Research,'' \textit{Biometrika}, vol. 82, no. 4, pp. 669–688, 1995.

% \bibitem{Pearl2009causality} J. Pearl, \textit{Causality}. Cambridge University Press, 2009.

% \bibitem{Pearl2012docalculus} J. Pearl, ``The Do-Calculus Revisited,'' \textit{arXiv preprint arXiv:1210.4852}, 2012.

\bibitem{RCA} K. B. Percarpio, V. B. Watts, and W. B. Weeks, ``The Effectiveness of Root Cause Analysis: What does the Literature Tell Us?'', \textit{The Joint Commission Journal on Quality and Patient Safety}, 2008.

\bibitem{Perez2022} E. Perez, S. Huang, F. Song, T. Cai, R. Ring, J. Aslanides, A. Glaese, N. McAleese, and G. Irving, ``Red Teaming Language Models with Language Models,'' \textit{arXiv preprint arXiv:2202.03286}, 2022.

% \bibitem{Peters2001} R. A. Peters, K. Hambuchen, K. Kawamura, and D. M. Wilkes, ``The Sensory Ego-Sphere as a Short-Term Memory for Humanoids,'' \textit{IEEE-RAS International Conference on Humanoid Robots}, 2001.

% \bibitem{Peters2009} R. A. Peters, K. A. Hambuchen, and R. E. Bodenheimer, ``The Sensory Ego-Sphere: A Mediating Interface Between Sensors and Cognition,'' \textit{Autonomous Robots}, 2009.

% \bibitem{Philippsen2009} R. Philippsen, N. Nejati, and L. Sentis, ``Bridging the Gap Between Semantic Planning and Continuous Control for Mobile Manipulation Using a Graph-Based World Representation,'' \textit{Workshop on Hybrid Control Autonomous Systems}, 2009.

% \bibitem{imetropicknik} PickNik, ``NASA JSC: Robotic Manipulation for Autonomous Lunar Operations at NASA,'' PickNik, 2023. [Online]. Available: \url{https://picknik.ai/case-study-nasa-jsc/}

\bibitem{MoveIt2_2024} PickNik Robotics, ``MoveIt2 Documentation,'' PickNik, 2024. [Online]. Available: \url{https://moveit.picknik.ai/main/index.html}

% \bibitem{Porcheron2018} M. Porcheron, J. E. Fischer, S. Reeves, and S. Sharples, ``Voice Interfaces in Everyday Life,'' \textit{CHI Conference on Human Factors in Computing Systems}, 2018.

% Q----------

% \bibitem{RL2011} W. Qiang and Z. Zhongli, ``Reinforcement Learning Model, Algorithms, and Its Application,'' \textit{International Conference on Mechatronic Science, Eletric Engineering, and Computer (MEC)}, 2011.

% R----------

\bibitem{Radford2015} N. A. Radford, P. Strawser, K. Hambuchen, J. S. Mehling, W. K. Verdeyen, A. S. Donnan, J. Holley, J. Sanchez, V. Nguyen, L. Bridgwater, R. Berka, R. Ambrose, M. M. Markee, and N. J. Fraser-Chanpong, ``Valkyrie: NASA's First Bipedal Humanoid Robot,'' \textit{Journal of Field Robotics}, 2015.

\bibitem{Randhawa2018} S. Randhawa, B. Turnbull, J. Yuen, and J. Dean, ``Mission-Centric Automated Cyber Red Teaming,'' \textit{International Conference on Availability, Reliability and Security}, 2018.

\bibitem{YOLOpaper} J. Redmon, S. Divvala, R. Girshick, and A. Farhadi, ``You Only Look Once: Unified, Real-Time Object Detection,'' \textit{IEEE Conference on Computer Vision and Pattern Recognition (CVPR)}, 2016.

\bibitem{Robinette2016} P. Robinette, W. Li, R. Allen, A. M. Howard, and A. R. Wagner, ``Overtrust of Robots in Emergency Evacuation Scenarios,'' \textit{IEEE International Conference on Human-Robot Interaction (HRI)}, 2016.

% \bibitem{DQN2017} M. Roderick, J. MacGlashan, and S. Tellex, ``Implementing the Deep Q-Network,'' \textit{arXiv preprint arXiv:1711.07478}, 2017.

% \bibitem{Roese1997} N. J. Roese, ``Counterfactual Thinking,'' \textit{Psychological Bulletin}, 1997.

\bibitem{Roese2017} N. J. Roese and K. Epstude, ``The Functional Theory of Counterfactual Thinking: New Evidence, New Challenges, New Insights,'' \textit{Advances in Experimental Social Psychology}, 2017.

% \bibitem{Fox2021} J. Roh, K. Desingh, A. Farhadi, and D. Fox, ``LanguageRefer: Spatial-Language Model for 3D Visual Grounding,'' \textit{Conference on Robot Learning (CoRL)}, 2021.

% \bibitem{SafetyHouseRepairs} D. Roos, ``10 Home repairs That Can Seriously Break the Bank,'' HowStuffWorks, 2014. [Online]. Available: \url{https://home.howstuffworks.com/home-improvement/repair/10-home-repairs-break-bank.htm}

% \bibitem{FrameNet2010} J. Ruppenhofer, M. Ellsworth, M. Schwarzer-Petruck, C. R. Johnson, and J. Scheffczyk, ``FrameNet II: Extended Theory and Practice,'' \textit{International Computer Science Institute}, 2016.

% \bibitem{RussellNorvig2010} S. J. Russell and P. Norvig, \textit{Artificial Intelligence: A Modern Approach, Third Edition}.  Pearson Education, 2010.

\bibitem{RussellNorvig2020} S. J. Russell and P. Norvig, \textit{Artificial Intelligence: A Modern Approach, Fourth Edition}. Pearson Education, 2020.

% S----------

\bibitem{Schneier2012} B. Schneier, ``Liars \& Outliers: Enabling the Trust that Society Needs to Thrive,'' \textit{John Wiley \& Sons}, 2012.

\bibitem{Schneier2015} B. Schneier, ``Secrets and Lies: Digital Security in a Networked World,'' \textit{John Wiley \& Sons}, 2015.

\bibitem{Schulte2020} P. A. Schulte, J. M. K. Streit, F. Sheriff, G. Delclos, S. A. Felknor, S. L. Tamers, S. Fendinger, J. Grosch, and R. Sala, ``Potential Scenarios and Hazards in the Work of the Future: A Systematic Review of the Peer-Reviewed and Gray Literatures,'' \textit{Annals of Work Exposures and Health}, 2020.

\bibitem{FMEA} K. D. Sharma and S. Srivastava, ``Failure Mode and Effect Analysis (FMEA) Implementation: A Literature Review,'' \textit{Journal of Advance Research in Aeronautics and Space Science}, 2018.

\bibitem{She2015} L. She, Y. Jia, N. Xi, and J. Y. Chai, ``Exception Handling for Natural Language Control of Robots,'' \textit{IEEE International Conference on Human-Robot Interaction Extended Abstracts}, 2015.

% \bibitem{sheetz2024hsfs} E. Sheetz, M. Shannon, C. Kisailus, A. Ingerman, and S. Azimi, ``Hierarchical Semantic Frames for Grounding Language in Robot Control Primitives,'' NASA Technical Reports Server (NTRS), 2024. [Online]. Available: \url{https://ntrs.nasa.gov/citations/20240004454}

\bibitem{Sheridan2016} T. B. Sheridan, ``Human-Robot Interaction: Status and Challenges,'' \textit{Human Factors}, 2016.

% \bibitem{Fox2022} M Shridhar, L. Manuelli, and D. Fox, ``Cliport: What and Where Pathways for Robotic Manipulation,'' \textit{Conference on Robot Learning (CoRL)}, 2022.

% \bibitem{DQN2013} D. Silver, A. Graves, I. Antonoglou, M. Riedmiller, V. Mnih, D. Wierstra, and K. Kavukcuoglu, ``Playing Atari with Deep Reinforcement Learning,'' \textit{DeepMind Lab. arXiv}, 2013.

% \bibitem{Storks2021} S. Storks, Q. Gao, Y. Zhang, and J. Chai, ``Tiered Reasoning for Intuitive Physics: Toward Verifiable Commonsense Language Understanding,'' \textit{arXiv preprint arXiv:2109.04947}, 2021.

% T----------

\bibitem{Tan2014} T. Tan, S. Porter, T. Tan, and G. West, ``Computational Red Teaming for Physical Security Assessment,'' \textit{IEEE International Conference on Cyber Technology in Automation, Control, and Intelligent Systems}, 2014.

% \bibitem{Tellex2011} S. Tellex, T. Kollar, S. Dickerson, M. R. Walter, A. G. Banerjee, S. Teller, and N. Roy, ``Understanding Natural Language Commands for Robotics Navigation and Mobile Manipulation,'' \textit{AAAI Conference on Artificial Intelligence}, 2011.

% \bibitem{Tellex2020} S. Tellex, N. Gopalan, H. Kress-Gazit, and C. Matuszek, ``Robots that Use Language,'' \textit{Annual Review of Control, Robotics, and Autonomous Systems}, 2020.

% \bibitem{SafetyVehicle} C. Threewitt, ``10 Serious Mistakes in Car Maintenance,'' HowStuffWorks, 2013. [Online]. Available: \url{https://auto.howstuffworks.com/under-the-hood/vehicle-maintenance/10-serious-mistakes-car-maintenance.htm}

% \bibitem{Thomas1997} J. Thomas, ``Conversational Maxims,'' \textit{Concise Encyclopedia of Philosophy of Language}, 1997.

% \bibitem{RoboFrameNet2012} B. J. Thomas and O. C. Jenkins, ``RoboFrameNet: Verb-Centric Semantics for Actions in Robot Middleware,'' \textit{IEEE International Conference on Robotics and Automation (ICRA)}, 2012.

% \bibitem{Tojo2000} T. Tojo, Y. Matsusaka, T. Ishii, and T. Kobayashi, ``A Conversational Robot Utilizing Facial and Body Expressions,'' \textit{IEEE International Conference on Systems, Man, and Cybernetics (SMC)}, 2000.

% \bibitem{Trandabat2010} D. Trandabat and D. Cristea, ``Natural Language Processing Using Semantic Frames,'' Ph.D. dissertation, University ``Alexandru Ioan Cuza'' of Ia\c{s}i, Romania, 2010.

% U----------

% \bibitem{FMEA_UT} University of Texas School of Public Health, ``Failure Mode Effects Analysis (FMEA),'' UTHealth, 2014. [Online]. Available: \url{https://www.slideserve.com/primo/failure-mode-effects-analysis-fmea-2601014}

% V----------

\bibitem{colorblob} J. Van De Weijer, C. Schmid, J. Verbeek, and D. Larlus, ``Learning Color Names for Real-World Applications,'' \textit{IEEE Transactions on Image Processing}, 2009.

% \bibitem{Vasic2013} M. Vasic and A. Billard, ``Safety Issues in Human-Robot Interactions,'' \textit{IEEE International Conference on Robotics and Automation (ICRA)}, 2013.

% W----------

% \bibitem{Wachter2017} S. Wachter, B. Mittelstadt, and C. Russell, ``Counterfactual Explanations without Opening the Black Box: Automated Decisions and the GDPR,'' \textit{Harv. JL \& Tech.}, 2017.

% \bibitem{Wang2011} Y. Y. Wang, L. Deng, and A. Acero, ``Semantic Frame-Based Spoken Language Understanding,'' \textit{Spoken Language Understanding: Systems for Extracting Semantic Information from Speech}, 2011.

% \bibitem{Winograd1971} T. Winograd, ``Procedures as a Representation for Data in a Computer Program for Understanding Natural Language,'' MIT Technical Report, 1971.

% \bibitem{SHRDLU1972} T. Winograd, ``SHRDLU: A System for Dialog,'' 1972.

\bibitem{Wood2000} B. J. Wood and R. A. Duggan, ``Red Teaming of Advanced Information Assurance Concepts,'' \textit{IEEE DARPA Information Survivability Conference and Exposition (DISCEX)}, 2000.

% X----------



% Y----------

\bibitem{Yang2006} A. Yang, H. A. Abbass, and R. Sarker, ``Characterizing Warfare in Red Teaming,'' \textit{IEEE Transactions on Systems, Man, and Cybernetics (Cybernetics)}, 2006.

% Z----------

\bibitem{Zacharaki2020} A. Zacharaki, I. Kostavelis, A. Gasteratos, and I. Dokas, ``Safety Bounds in Human Robot Interaction: A Survey,'' \textit{Safety Science}, 2020.

% \bibitem{Zech2017} P. Zech, S. Haller, S. R. Lakani, B. Ridge, E. Ugur, and J. Piater, ``Computational Models of Affordance in Robotics: A Taxonomy and Systematic Classification,'' \textit{Adaptive Behavior}, 2017.

\bibitem{Zenko2015} M. Zenko, \textit{Red Team: How to Succeed by Thinking Like the Enemy}, Basic Books, 2015.

\bibitem{Zhang2020} J. Zhang and W. Song, ``Physics-of-Failure Based Model for Industrial Robot Reliability Prediction,'' \textit{IEEE International Conference on Mechatronics and Automation (ICMA)}, 2020.

\bibitem{Zhang2022} Y. Zhang, J. Yang, J. Pan, S. Storks, N. Devraj, Z. Ma, K. P. Yu, Y. Bao, and J. Chai, ``DANLI: Deliberative Agent for Following Natural Language Instructions,'' \textit{arXiv preprint arXiv:2210.12485}, 2022.

% \bibitem{Ziegler2015} L. Ziegler, ``The Attentive Robot Companion: Learning Spatial Information from Observation and Verbal Interaction,'' Ph.D. Dissertation, Bielefeld University, 2015.

\end{thebibliography}
\end{document}